%% file: main.tex
\documentclass[final,5p,times,twocolumn]{elsarticle}

\input{0a-packages}
\input{0b-config}

\journal{Neurocomputing}

\begin{document}

\begin{frontmatter}


\title{NeuroTrain: Surveying Local Learning Rules for Spiking Neural Networks with an Open Benchmarking Framework 
}

\author[polito]{Alessio Caviglia\corref{cor1}}
\ead{alessio.caviglia@polito.it}
\author[polito]{Filippo Marostica}
\ead{filippo.marostica@polito.it}
\author[polito]{Roberta Bardini}
\ead{roberta.bardini@polito.it}
\author[polito]{Alessandro Savino}
\ead{alessandro.savino@polito.it}
\author[polito]{Stefano Di Carlo}
\ead{stefano.dicarlo@polito.it}
\cortext[cor1]{Corresponding author}

\affiliation[polito]{organization={Politecnico di Torino, Control and Computer Engineering Department},
            addressline={Corso Duca degli Abruzzi 24}, 
            city={Torino},
            postcode={I-10129}, 
            state={},
            country={Italy}}

\input{0c-abstract}




\end{frontmatter}



\input{1-introduction}

\input{2-background-new}
\input{3-related_works}
\input{4-taxonomy}
\input{5-software}

\input{6-conclusion}

\bibliographystyle{elsarticle-num} 
\bibliography{references}

\end{document}

%% file: 0a-packages.tex
\usepackage{amsmath,amssymb}

\usepackage{booktabs}
\usepackage{multirow}
\usepackage{array}

\usepackage[hidelinks]{hyperref}

\usepackage[acronym]{glossaries}
\makeglossaries

\usepackage[table]{xcolor}

\usepackage{rotating}

%% file: 0b-config.tex
\glsdisablehyper
\biboptions{sort&compress}

\newcommand{\x}{\textbf{X}}
\newcommand{\m}{\ensuremath{\circ}} 

\newacronym{adse}{ADSE}{Automatic Design Space Exploration}
\newacronym{aer}{AER}{Address Event Representation}
\newacronym{ai}{AI}{Artificial Intelligence}
\newacronym{ann}{ANN}{Artificial Neural Network}
\newacronym{api}{API}{Application Programming Interface}
\newacronym{ax}{AX}{Adaptive eXperimentation}
\newacronym{bptt}{BPTT}{Back-Propagation Through Time}
\newacronym{bram}{BRAM}{Block RAM}
\newacronym{cl}{CL}{Continual Learning}
\newacronym{cnn}{CNN}{Convolutional Neural Network}
\newacronym{cots}{COTS}{Commercial Off The Shelf}
\newacronym{cpu}{CPU}{Central Processing Unit}
\newacronym{csnn}{CSNN}{Convolutional Spiking Neural Network}
\newacronym{cu}{CU}{Control Unit}
\newacronym{dnn}{DNN}{Deep Neural Network}
\newacronym{dp}{DP}{Data Path}
\newacronym{dram}{DRAM}{Dynamic Random Access Memory}
\newacronym{dse}{DSE}{Design Space Exploration}
\newacronym{dvs}{DVS}{Dynamic Vision Sensor}
\newacronym{eda}{EDA}{Electronic Design Automation}
\newacronym{elut}{ELUT}{Equivalent Look Up Table}
\newacronym{eth}{ETH}{Eidgenössische Technische Hochschule}
\newacronym{fc}{FC}{Fully-Connected}
\newacronym{fcr}{FC-R}{Fully-Connected Recurrent}
\newacronym{ff}{FF}{Forward-Forward}
\newacronym{fffc}{FF-FC}{Feed-Forward Fully-Connected}
\newacronym{FFNN}{FFNN}{Feed-Forward Neural Network}
\newacronym{fi}{FI}{Fault Injection}
\newacronym{fim}{FIM}{Fault Injection Manager}
\newacronym{fl}{FL}{Fault List}
\newacronym{flg}{FLG}{Fault List Generator}
\newacronym{fpga}{FPGA}{Field Programmable Gate Array}
\newacronym{fsm}{FSM}{Finite State Machine}
\newacronym{gpgpu}{GPGPU}{General Purpose Graphic Processing Unit}
\newacronym{gpu}{GPU}{Graphic Processing Unit}
\newacronym{hdl}{HDL}{Hardware Description Language}
\newacronym{hw}{HW}{Hardware}
\newacronym{if}{IF}{Integrate and Fire}
\newacronym{ini}{INI}{Institute of Neuro-Informatics}
\newacronym{iot}{IoT}{Internet of Things}
\newacronym{lif}{LIF}{Leaky Integrate and Fire}
\newacronym{lr}{LR}{Latent Replay}
\newacronym{lstm}{LSTM}{Long Short Term Memory}
\newacronym{lut}{LUT}{Look Up Table}
\newacronym{mac}{MAC}{Multiply and Accumulate}
\newacronym{ml}{ML}{Machine Learning}
\newacronym{mlp}{MLP}{Multi-Layer Perceptron}
\newacronym{nas}{NAS}{Network Architecture Search}
\newacronym{ne}{NE}{Network Evaluator}
\newacronym{ng}{NG}{Network Generator}
\newacronym{nir}{NIR}{Neuromorphic Intermediate Representation}
\newacronym{nlp}{NLP}{Natural Language Processing}
\newacronym{nn}{NN}{Neural Network}
\newacronym{ostl}{OSTL}{Online Spatio-Temporal Learning}
\newacronym{ostp}{OSTP}{Online Spatio-Temporal Learning with Target Projection}
\newacronym{pu}{PU}{Processing Unit}
\newacronym{pulp}{PULP}{Parallel processing Ultra-Low Power}
\newacronym{qat}{QAT}{Quantization Aware Training}
\newacronym{ram}{RAM}{Random Access Memory}
\newacronym{rcr}{RC-R}{Randomly-Connected Recurrent}
\newacronym{rl}{RL}{Reinforcement Learning}
\newacronym{rnn}{RNN}{Recurrent Neural Network}
\newacronym{rom}{ROM}{Read Only Memory}
\newacronym{rsnn}{RSNN}{Recurrent Spiking Neural Network}
\newacronym{rtl}{RTL}{Register Transfer Level}
\newacronym{rtrl}{RTRL}{Real-Time Recurrent Learning}
\newacronym{sbs}{SbS}{Spike-by-Spike}
\newacronym{scnn}{SCNN}{Spiking Convolutional Neural Networks}
\newacronym{sfi}{SFI}{Statistical Fault Injection}
\newacronym{sg}{SG}{Surrogate Gradient}
\newacronym{shd}{SHD}{Spiking Heidelberg Dataset}
\newacronym{simd}{SIMD}{Single Instruction Multiple Data}
\newacronym{snn}{SNN}{Spiking Neural Network}
\newacronym{soc}{SoC}{System on Chip}
\newacronym{sota}{SOTA}{State Of The Art}
\newacronym{sram}{SRAM}{Static Random Access Memory}
\newacronym{srm}{SRM}{Spike Response Model}
\newacronym{stdp}{STDP}{Spike-Timing-Dependent Plasticity}
\newacronym{tpu}{TPU}{Tensor Processing Unit}
\newacronym{ucb}{UCB}{Upper Confidence Bound}
\newacronym{vhdl}{VHDL}{VHSIC Hardware Description Language}
\newacronym{wta}{WTA}{Winner Takes All}
\newacronym{PNN}{PNN}{Photonic Neural Network}
\newacronym{ADC}{ADC}{Analog-to-Digital Converter}
\newacronym{DAC}{DAC}{Digital-to-Analog Converter}
\newacronym{MZI}{MZI}{Mach-Zehnder Interferometer}
\newacronym{MVM}{MVM}{Matrix-Vector Multiplication}
\newacronym{CMOS}{CMOS}{Complementary Metal-Oxide-Semiconductor}
\newacronym{asic}{ASIC}{Application-Specific Integrated Circuit}
\newacronym{SVD}{SVD}{Singular Value Decomposition}
\newacronym{epsp}{EPSP}{Excitatory Post-Synaptic Potential}
\newacronym{ltp}{LTP}{Long-Term Potentiation}
\newacronym{ltd}{LTD}{Long-Term Depression}
\newacronym{ptp}{PTP}{Post-Tetanic Potentiation}
\newacronym{bp}{BP}{Back-Propagation}
\newacronym{shl}{SHL}{Supervised Hebbian Learning}
\newacronym{stbp}{STBP}{Spatio-Temporal Back-Propagation}

\newacronym{dfa}{DFA}{Direct Feedback Alignment}
\newacronym{stsf}{STSF}{Spiking Time Sparse Feedback}
\newacronym{bcm}{BCM}{Bienenstock--Cooper--Munro}
\newacronym{bell}{BELL}{Backward Exact Local Learning}
\newacronym{clapp}{CLAPP}{Contrastive Local And Predictive Plasticity}
\newacronym{decolle}{DECOLLE}{Deep Continuous Local Learning}
\newacronym{drtp}{DRTP}{Direct Random Target Projection}
\newacronym{eicil}{EICIL}{Excitatory Inhibitory Cycle Iteration Learning}
\newacronym{ell}{ELL}{Exact Local Learning}
\newacronym{eprop}{e-prop}{Eligibility Propagation}
\newacronym{etlp}{ETLP}{Event-based Three-factor Local Plasticity}
\newacronym{fell}{FELL}{Forward Exact Local Learning}
\newacronym{osttp}{OSTTP}{Online Spatio-Temporal Learning with Target Projection}
\newacronym{ottt}{OTTT}{Online Training Through Time}
\newacronym{resume}{ReSuMe}{Remote Supervised Method}
\newacronym{rstdp}{R-STDP}{Reward-modulated Spike-Timing-Dependent Plasticity}
\newacronym{s-tllr}{S-TLLR}{STDP-inspired Temporal Local Learning Rule}
\newacronym{sltt}{SLTT}{Spatial Learning Through Time}
\newacronym{spsp}{SPSP}{Spike-Train-level Post-Synaptic Potential}
\newacronym{stdfa}{ST-DFA}{Spike-Train-level Direct Feedback Alignment}
\newacronym{tess}{TESS}{Temporally and Spatially Local Learning Rule}
\newacronym{sdsp}{SDSP}{Spike-Driven Synaptic Plasticity}
\newacronym{popsan}{PopSAN}{Population-coded Spiking Actor Network}
\newacronym{fptt}{FPTT}{Forward Propagation Through Time}
\newacronym{tp}{TP}{Traces Propagation}
\newacronym{mcmc}{MCMC}{Markov Chain Monte Carlo}
\newacronym{otpe}{OTPE}{Online Training with Postsynaptic Estimates}
\newacronym{csdp}{CSDP}{Contrastive Signal-Dependent Plasticity}

%% file: 0c-abstract.tex
\begin{abstract}
The rapid expansion of spiking neural networks (SNNs) has led to a proliferation of training algorithms that differ widely in biological inspiration, computational structure, and hardware suitability. Despite this progress, the field lacks a unified, fine-grained taxonomy that systematically organizes these approaches and clarifies their conceptual relationships. This survey provides a comprehensive taxonomy of SNN training algorithms, spanning surrogate-gradient backpropagation, local and three-factor learning rules, biologically inspired plasticity mechanisms, ANN-to-SNN conversion pipelines, and non-standard optimization strategies. We analyze each class in terms of its computational principles, learning signals, and locality properties. To support reproducible research, we release NeuroTrain, an open-source snnTorch-based framework that implements a representative set of these algorithms within a unified, modular, and extendable framework, enabling consistent benchmarking across datasets, architectures, and training regimes. By consolidating fragmented literature and providing a reusable benchmarking framework, this survey identifies common patterns, highlights open challenges, and outlines promising directions for future work on scalable, efficient SNN training.
\end{abstract}

%% file: 1-introduction.tex
\section{Introduction}
\label{sec:intro}

\Glspl{snn} are event-driven neural networks in which information is represented and processed through spikes~\cite{maass_networks_1997}. \glspl{snn} incorporate temporal dynamics and can exploit sparse activity, i.e., computation is triggered only when events occur. This enables low-latency processing and the potential for substantial energy savings when deployed on neuromorphic hardware \cite{davies_loihi_2018}. These properties make \glspl{snn} attractive for edge and always-on sensing scenarios, where power, bandwidth, and response time are often the dominant constraints. Beyond their relevance to neuromorphic hardware, they also provide a theoretically grounded setting for studying computation under constraints that closely resemble those of biological neural systems \cite{huo_research_2025}.

The main obstacle to widespread adoption of \glspl{snn}, however, remains \emph{training}: while forward simulation is straightforward, assigning credit for errors through discontinuous spike events and temporal dynamics is substantially more complex than in other \glspl{ann}.
Training \glspl{snn} is challenging for at least three intertwined reasons. First, spike generation is non-differentiable, making gradient-based optimization non-trivial and requiring approximations (e.g., surrogate gradients). Second, information is carried over time through membrane and synaptic state variables, so learning must address temporal credit assignment across many time steps. Third,  implementations face substantial computational and memory costs: methods that rely on global backpropagation through time can be accurate but expensive, and they map poorly onto distributed neuromorphic hardware where centralized memory access and global error transport are undesirable.
These difficulties have catalyzed a large and rapidly growing body of work on \gls{snn} learning, spanning direct training, conversion pipelines, and alternative optimization strategies. This survey focuses on \emph{direct training} methods,  which train \glspl{snn} by explicitly modeling their temporal dynamics and spike-generation processes during optimization, rather than relying on \gls{ann} to \gls{snn} conversion.

Within direct training, \emph{locality} is often highlighted as a defining characteristic. Broadly, locality concerns the availability of the information required to compute learning updates: \emph{temporal} locality refers to updates that can be computed from quantities available at the current timestep. In contrast, \emph{spatial} locality refers to updates that depend on signals available locally at the synapse. Locality reduces global communication and memory pressure and aligns well with low-power neuromorphic hardware and online adaptation, yet relaxing locality often improves training performance on challenging benchmarks. A key goal of this survey is to discuss these trade-offs and provide instruments for comparison across methods.

Several surveys in recent years have provided valuable overviews of \gls{snn} training, spanning direct training methods, conversion pipelines, and other approaches (see \autoref{sec:related_work}). However, the pace at which new algorithms appear makes the landscape difficult to track, and, perhaps more importantly, \emph{fair comparison remains hard}. Existing surveys are primarily descriptive: they synthesize ideas and trends, but typically cannot provide standardized, side-by-side evaluations because methods live in heterogeneous codebases with differing datasets, preprocessing pipelines, architectures, optimization details, and reporting conventions. As a result, it is often difficult to disentangle algorithmic contributions from experimental choices, and readers may come away with a rich qualitative map of approaches but a weaker sense of \emph{which} design decisions matter most under controlled, unified conditions.

In this work, we address this gap by considering the \emph{benchmarking} as a key contribution, paired with a taxonomy that organizes the literature along explicit axes of variation, i.e., \emph{training strategy}, \emph{supervision}, and \emph{locality}. While we cover the major training paradigms, we place particular emphasis on direct training, where the design space is richest, and locality constraints most directly shape both algorithmic structure and hardware suitability. Concretely, our contributions are twofold: (i) a unified taxonomy-driven review of \gls{snn} training methods; and (ii) \emph{NeuroTrain}, an open, extensible experimental framework that enables reproducible implementations and comparisons of \gls{snn} training algorithms under shared experimental conditions. Beyond enabling replication, \emph{NeuroTrain} is designed to lower the barrier to integrating new algorithms by providing a common suite of reference models and pipelines for testing them. In this sense, the present work goes beyond a conventional survey: it introduces reusable benchmarking tools to support more rigorous, cumulative progress in this rapidly evolving field.

The remainder of the article is structured as follows. Section~\ref{sec:background} introduces the background required to understand the \gls{snn} training challenges. Section~\ref{sec:related_work} reviews related surveys on \glspl{snn} training. Section~\ref{sec:taxonomy} presents the proposed taxonomy. Section~\ref{sec:software_framework} describes the benchmarking framework, experimental design, and results. Finally, Section~\ref{sec:conclusion} concludes the article and outlines future research directions.

%% file: 2-background-new.tex
\section{Background}
\label{sec:background}

\glspl{snn} are a class of neural models in which information is represented and processed through discrete spikes evolving over continuous or discretized time \cite{maass_networks_1997}. By explicitly modeling neuronal dynamics over time, \glspl{snn} differ fundamentally from conventional \glspl{ann}, which generally abstract neurons as static input–output units rather than temporally evolving, spike-emitting systems. This event-driven formulation has made \glspl{snn} particularly relevant to neuromorphic computing, where sparse activity and asynchronous computation can be exploited to achieve low-latency, energy-efficient processing. At the same time, these same properties make \glspl{snn} substantially more difficult to train. 
This section establishes the minimal conceptual framework for understanding the challenges of \gls{snn} training.

\subsection{Spiking computation and neuronal dynamics}

\begin{figure}[!ht]
    \centering
    \includegraphics[width=0.99\linewidth]{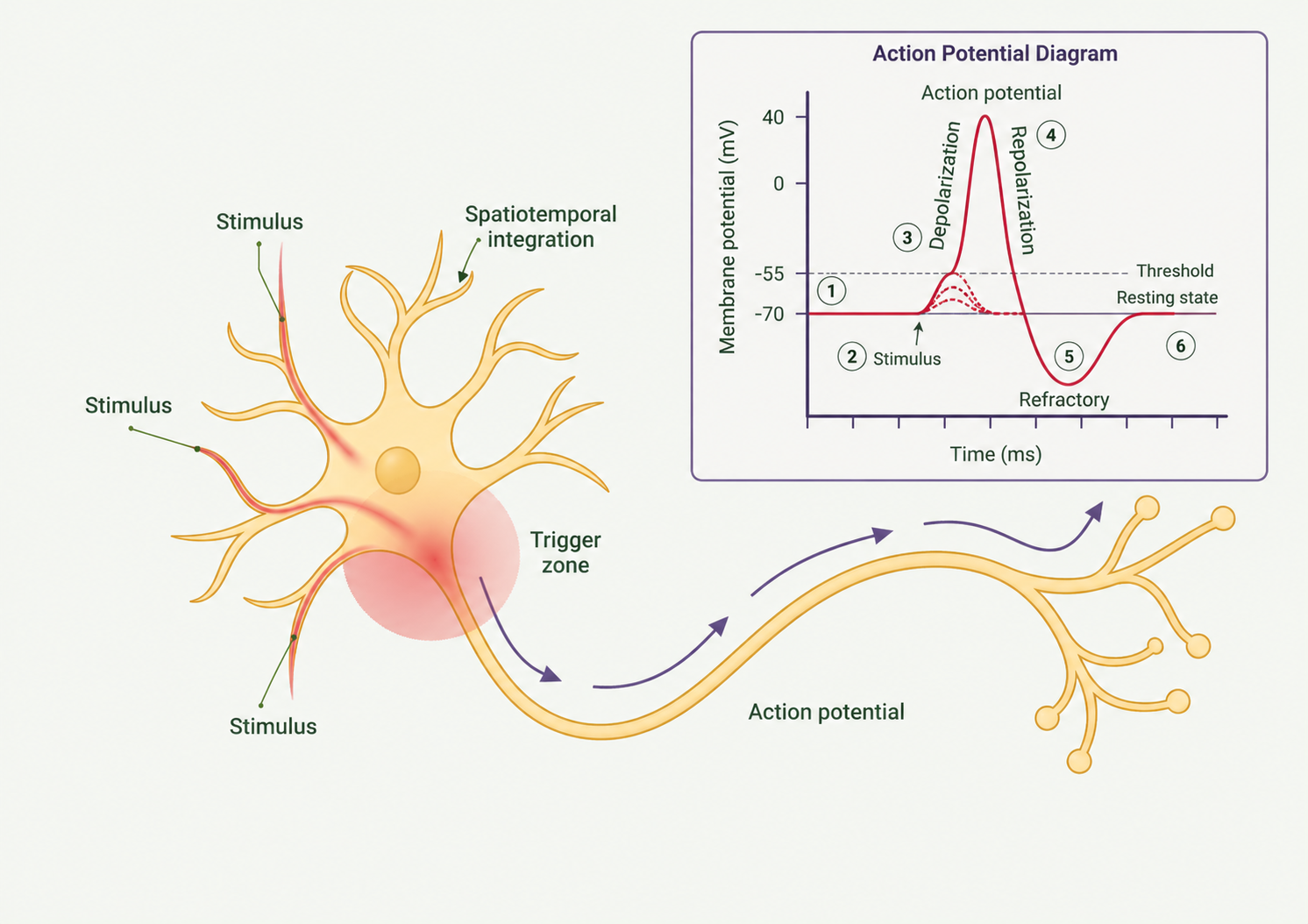}
    \caption{Biological neuron behavior. Spatiotemporal integration and action potential dynamics at the single neuron.}
    \label{fig:ap}
\end{figure}

Biological neurons communicate through action potentials and integrate synaptic inputs over time before producing an output spike (\autoref{fig:ap}). Incoming signals from different \emph{locations} arriving at different \emph{times} are accumulated into an evolving membrane potential, and a spike is emitted once this state crosses a threshold, typically followed by reset and refractoriness. Spiking neuron models used in \glspl{snn}, such as the \acrfull{lif} model, abstract these mechanisms into low-dimensional dynamical systems that preserve the key computational ingredients of spatio-temporal integration, thresholded event generation, and post-spike reset \cite{lansky2008review,huo_research_2025}.

This formulation endows \glspl{snn} with a computational regime in which information may be carried not only by activation magnitude, but also by spike timing and temporal structure. Such temporal expressivity is one of the principal motivations for using \glspl{snn}, especially in domains involving streams, events, or dynamical signals. However, it also introduces a central optimization difficulty that impedes training these networks. 

\subsection{Why training \glspl{snn} is challenging}

The training of \glspl{snn} is difficult for several interconnected reasons. First, the binary and threshold-based nature of spike emission prevents the direct application of standard gradient-based optimization in its usual form~\cite{guo_direct_2023}. Moreover, because \glspl{snn} are dynamical systems, parameter updates must often account for how present outputs depend on earlier hidden states and past inputs~\cite{meszarosEfficientEventbasedDelay2025}. Eventually, the sparse, event-driven nature of spiking activity can make optimization unstable or inefficient, particularly when useful supervisory information is weak, delayed, or distributed over time~\cite{xiaoSPIDEPurelySpikebased2023}.

These challenges have given rise to a broad range of learning strategies. Some methods avoid direct optimization of spiking dynamics by transferring knowledge from rate-based or non-spiking models. In contrast, others train spiking networks directly by explicitly accounting for their temporal behavior. Within this landscape, a particularly important distinction concerns the nature of the information required to compute synaptic updates. Some methods rely on global error signals propagated across layers and over time, whereas others construct updates using only locally available signals within the network. The latter class is especially relevant to this survey.

\subsection{Locality as a principle of biological plasticity}

A central feature of biological learning is that synaptic and neuronal changes are driven predominantly by local variables. In the nervous system, plasticity is generally understood to arise from activity-dependent modifications that depend on information available at or near the site of change, including presynaptic activity, postsynaptic activity, membrane state, local biochemical traces, and modulatory influences. This spatial and temporal locality is one of the main reasons local learning rules are commonly regarded as more biologically grounded than algorithms based on exact end-to-end error backpropagation.

The classical conceptual basis for local adaptation is Hebbian plasticity, which holds that coordinated activity between neurons tends to strengthen their functional coupling. In spiking systems, this principle is refined into \gls{stdp}, where synaptic modification depends on the relative timing of pre- and postsynaptic spikes \cite{bi2002spatiotemporal}. When presynaptic spikes reliably precede postsynaptic spikes within an appropriate temporal window, potentiation may occur; when the order is reversed, depression may result. \acrshort{stdp} thus provides a canonical example of a rule that is local both in space, because it depends on synapse-specific activity, and in time, because it depends on temporally precise spike relationships.
Importantly, biological plasticity cannot be reduced to a single timing-based mechanism. Experimental evidence indicates that learning emerges from the interaction of multiple processes, including synaptic plasticity, intrinsic plasticity, homeostatic regulation, and neuromodulator-dependent modulation \cite{zha2022molecular,sweatt2016neural,mcfarlan2023plasticitome}. These mechanisms operate across multiple timescales and cellular compartments, and their coordinated action contributes to stability, adaptability, and functional specialization. For the study of \glspl{snn}, this broader perspective is significant because many local learning algorithms can be understood as abstractions of such mechanisms, even when they do not aim to reproduce them literally.

\subsection{Local learning in \glspl{snn}}

In the context of this paper, \emph{local learning} refers to training schemes in which parameter updates are computed from information available locally at the synapse, neuron, or layer level, rather than from exact global gradients propagated throughout the entire network. The relevant local signals may include spike traces, membrane potentials, eligibility traces, local errors, lateral interactions, or modulatory factors, depending on the specific method under consideration.

This notion of locality should be interpreted as a principled algorithmic constraint rather than as a claim of strict biological realism. Most local training methods in machine learning are not intended as faithful models of cortical plasticity. Rather, they are motivated by the observation that biological learning appears to proceed without exact weight transport or globally coordinated backward passes, and that useful learning can instead emerge from the interaction of local activity, circuit structure, and modulatory processes. From this standpoint, local learning in \glspl{snn} occupies an intermediate position between neuroscience-inspired modeling and practical algorithm design.

\subsection{Relevance to neuromorphic computing}

The appeal of local learning is not only biological but also architectural. Neuromorphic hardware substrates are typically characterized by distributed computation, co-localized memory and processing, event-driven execution, and limited support for globally synchronized communication~\cite{ivanov_neuromorphic_2022}. Under such constraints, training rules that require long-range error propagation, exact symmetry between forward and backward pathways, or repeated global coordination may be difficult to implement efficiently, particularly in online or on-chip settings.

By contrast, learning rules based on locally accessible variables are better compatible with the distributed and asynchronous organization of neuromorphic systems \cite{caviglia2025sfattispikingfpgaaccelerator, mayr_spinnaker_2019, davies_loihi_2018, carpegna_spiker_2024, akopyan_truenorth_2015, benjamin_neurogrid_2014, marostica2025energyefficientdigitaldesigncomparative}. For this reason, locality has become a key organizing principle in the design of training algorithms for \glspl{snn}, linking biological plausibility, hardware efficiency, and algorithmic scalability. This convergence of motivations explains why local learning occupies a central place in contemporary research on spiking neural networks. 

\subsection{Implications for this survey}

The following sections build on this background to examine how training algorithms for \glspl{snn} navigate the trade-offs among optimization effectiveness, biological plausibility, and hardware realizability. Particular emphasis is placed on local learning methods, not as a single algorithmic family, but as a broad design space defined by the use of spatially and temporally local learning signals to train spiking systems.

%% file: 3-related_works.tex
\section{Related Work}
\label{sec:related_work}

The rise in popularity of \glspl{snn} over the last decades has been accompanied by survey and perspective works offering periodic wrap-ups of the state of the art and recent advancements. Given the breadth of concepts and techniques encompassed by \glspl{snn}, these works span architectures, encoding techniques, datasets, applications, and related topics. In this section, we analyze these surveys with the specific aim of identifying those that cover training \glspl{snn}. We comment on them to highlight their strengths and clarify their differences, and to motivate the choices behind our survey taxonomy and the accompanying benchmarking framework.
To position this work, we first provide a concise, chronological account of how training has been covered as the field has evolved, and then discuss more recent surveys by scope, highlighting where the current view of training algorithms remains fragmented.

\paragraph{Chronology-driven filtering of prior surveys}
When working in rapidly evolving fields such as neuromorphic computing and the training of \glspl{snn}, time becomes a critical factor in evaluating review papers. Earlier surveys, even if well-written and thorough, can quickly fall out of scope as the literature expands. At the same time, they are valuable to understand how the community's focus evolved, which techniques were emphasized, and which directions matured into today's dominant practices. In the following, we summarize representative works excluded from the present survey due to their temporal coverage and the methods they employ.
One of the first works in this category by Ponulak et al. \cite{ponulak_introduction_2011} provides a clear snapshot of the field in the early 2010s. Although positioned as a general introduction covering models, coding, and applications, it devotes substantial space to learning, surveying \gls{stdp}-like rules and early supervised methods such as ReSuMe \cite{ponulak_supervised_2010} and SpikeProp \cite{bohte_error-backpropagation_2002}, as well as reinforcement formulations. Toward the end of the previous decade, surveys became more frequent. Beyond differences in focus, several recurring themes emerged in this period: training by conversion from \glspl{ann} was widely discussed as a practical route to high accuracy, while direct training remained actively explored, with early algorithms receiving most attention and initial adaptations of backpropagation, together with surrogate gradient-like approaches, appearing in the survey landscape \cite{pfeiffer_deep_2018, tavanaei_deep_2019, wang_supervised_2020}. In parallel, local learning was often framed in terms of \gls{stdp} and its variants, often in an unsupervised setting \cite{vigneron_critical_2020, taherkhani_review_2020}. Several reviews from this timeframe are not centered on training, but still contain curated and informative sections on learning and optimization \cite{roy_towards_2019, dora_spiking_2021, nguyen_review_2021, christensen_2022_2022, ivanov_neuromorphic_2022, nunes_spiking_2022, schuman_opportunities_2022, 9782767, yamazaki_spiking_2022}.
After this phase, the emphasis of surveys shifted. As assessed in \cite{guo_direct_2023}, backpropagation techniques with surrogate gradients, particularly \gls{bptt} adaptations to \glspl{snn}, increasingly dominated as leading approaches for direct training on complex architectures and datasets. At the same time, local learning was increasingly discussed as a supervised candidate, beyond its established role in unsupervised settings, aligning with the broader goal of bringing machine-learning capabilities to constrained and low-power devices where backpropagation can be challenging. In this evolving landscape, \gls{stdp} remained a widely used unsupervised algorithm and continued to inspire new techniques \cite{khacef_spike-based_2023}. In parallel, more systematic meta-analyses began to appear, for instance in \cite{pietrzak_overview_2023}, which compares training approaches through a complexity-oriented lens.
As interest in \glspl{snn} and their training continued to grow, more structured, field-specific, and application-driven reviews emerged, including reviews focused on hardware platforms \cite{rathi_exploring_2023, al_abdul_wahid_survey_2024}. In particular, Frenkel et al. \cite{frenkel_bottom-up_2023} survey neuromorphic computing with a focus on hardware design, including a detailed section on the current state of the art in training. A broad training coverage spanning datasets and applications is presented in \cite{yiLearningRulesSpiking2023}; however, given its publication date, it does not include the latest proposed techniques. As \glspl{snn} extend toward more complex architectures, dedicated works report that top results in these settings are still largely associated with backpropagation-based training and conversion methods, which can achieve high accuracy in complex scenarios \cite{dampfhoffer_backpropagation-based_2024}.
Overall, this chronological view clarifies how the emphasis in \gls{snn} training has shifted over time, from early \gls{stdp}-inspired rules and classical supervised methods to conversion pipelines and, more recently, surrogate-gradient backpropagation, with renewed attention to local and online learning as a practical direction for constrained deployments. At the same time, it highlights how quickly surveys become outdated in this area, motivating an updated synthesis that consolidates the training landscape under a unified view.

\paragraph{Scope-driven positioning of recent surveys}
Complementary to the chronological view, we now analyze the landscape of more recent works by scope and focus. The goal is to clarify what each review covers well and what remains underdeveloped with respect to this work, which aims to provide a thorough taxonomy of training algorithms with an emphasis on local, direct training rules and practical benchmarking.

A broad overview of learning and training can be found in works such as \cite{khan_spiking_2025, ayasi_practical_2025, wu_review_2024}. These works survey \glspl{snn} more generally; consequently, they do not aim to provide a complete and systematic taxonomy of training algorithms, and training is necessarily treated as one component within a wider perspective.
Among works that lean more explicitly towards training, each focuses on a different aspect, as expected in a field that is varied and complex. Works such as \cite{hu_toward_2024, zhou_direct_2024, han_progress_2025} focus on large networks for complex problems, where backpropagation and conversion training dominate, and therefore give comparatively less attention to approaches more suited to embedded and resource-constrained deployments, including many bio-inspired algorithms. Other specialized works include \cite{fatima_minhas_continual_2025, mishra_survey_2023} focusing on continual learning, \cite{shen_evolutionary_2024} concentrating on evolutionary approaches, and \cite{karamimanesh_spiking_2025} analyzing FPGA hardware adaptations.
We also found works centered on biologically plausible and local learning rules, each focusing on a particular subspace. For instance, \cite{mazurekThreeFactorLearningSpiking2025} focuses on algorithms structured as three-factor learning rules. In contrast, \cite{huo_research_2025} emphasizes biological plausibility and local learning mechanisms, with a strong focus on the \gls{stdp}-family and related synaptic plasticity rules.
Finally, a major gap in the current landscape is the lack of clear, reproducible benchmarking. While work exists that compares algorithms primarily in terms of accuracy \cite{lin_benchmarking_2025}, these efforts typically consider only a limited subset of training rules and do not provide a practical, extensible toolchain for cross-method evaluation. Related software ecosystems (e.g., snnTorch \cite{eshraghian_training_2023}) facilitate implementation and experimentation, but, by themselves, they do not provide a survey-driven taxonomy coupled with systematic, reproducible comparisons across training approaches. A complementary perspective is taken by \cite{cheng_comprehensive_2025}, which benchmarks five neuromorphic training frameworks across image, text, and neuromorphic datasets under both direct training and \gls{ann}-to-\gls{snn} conversion, providing actionable guidance on framework selection but without covering local learning rules or supplying a taxonomy of training algorithms.

In summary, existing surveys collectively cover a broad range of \gls{snn} topics. Still, a consolidated view of training algorithms, particularly local methods, remains fragmented, and systematic comparisons across methods are still scarce. Table~\ref{tab:surveys} provides a compact map of the survey landscape, which we use to position the scope of this work. Building on these observations, the next sections introduce our taxonomy and benchmarking framework, with the explicit goal of enabling consistent, reproducible comparisons across training approaches.

\begin{table*}[tp]
  \centering
  \small
  \setlength{\tabcolsep}{8pt}
  \renewcommand{\arraystretch}{1.12}
  \begin{tabular}{l c >{\raggedright\arraybackslash}p{2.5cm}
                  c c c
                  c c c
                  c c}
    \toprule
    \multirow{2}{*}{\textbf{Work}}
      & \multirow{2}{*}{\textbf{Year}}
      & \multirow{2}{*}{\textbf{Focus}}
      & \multicolumn{3}{c}{\textbf{Training}}
      & \multicolumn{3}{c}{\textbf{Scope}}
      & \multicolumn{2}{c}{\textbf{Methodology}} \\
    \cmidrule(lr){4-6}\cmidrule(lr){7-9}\cmidrule(l){10-11}
      & & &
      \textbf{Conv} & \textbf{BP} & \textbf{Loc} &
      \textbf{Train} & \textbf{Large} & \textbf{HW} &
      \textbf{Taxon} & \textbf{Bench} \\
    \midrule

    \multicolumn{11}{l}{\textit{Chronology-driven set}} \\
    \midrule

    \cite{ponulak_introduction_2011}
      & 2011 & Introduction
      &       & \m   & \x
      & \x   &      & \m
      & \m   &      \\

    \cite{pfeiffer_deep_2018}
      & 2018 & Training survey
      & \m   & \x   & \m
      & \x   & \m   & \x
      & \x   &      \\

    \cite{tavanaei_deep_2019}
      & 2019 & Training survey
      & \x   & \x   & \m
      & \x   & \x   & \m
      & \m   &      \\

    \cite{roy_towards_2019}
      & 2019 & Perspective
      & \m   & \m   & \m
      & \m   &      & \x
      & \m   &      \\

    \cite{wang_supervised_2020}
      & 2020 & Training survey
      & \m   & \x   & \m
      & \x   & \m   &
      & \x   & \m   \\

    \cite{vigneron_critical_2020}
      & 2020 & Local learning
      &      &      & \x
      & \x   &      &
      & \m   &      \\

    \cite{taherkhani_review_2020}
      & 2020 & Local learning
      & \x   & \x   & \x
      & \x   &      & \m
      & \m   &      \\

    \cite{dora_spiking_2021}
      & 2021 & General SNN
      & \m   & \m   & \m
      & \m   & \m   & \m
      &      &      \\

    \cite{nguyen_review_2021}
      & 2021 & HW design
      & \m   & \m   & \m
      & \m   & \m   & \x
      &      &      \\

    \cite{christensen_2022_2022}
      & 2022 & Perspective
      &      &      & \m
      &      &      & \x
      &      &      \\

    \cite{ivanov_neuromorphic_2022}
      & 2022 & HW design
      &      &      & \m
      &      &      & \x
      &      &      \\

    \cite{nunes_spiking_2022}
      & 2022 & General SNN
      & \m   & \m   & \m
      & \m   & \m   & \m
      & \m   &      \\

    \cite{schuman_opportunities_2022}
      & 2022 & Perspective
      & \m   & \m   & \m
      &      &      & \m
      &      &      \\

    \cite{9782767}
      & 2022 & HW design
      &      &      & \m
      &      &      & \x
      &      &      \\

    \cite{yamazaki_spiking_2022}
      & 2022 & General SNN
      & \m   & \m   & \m
      & \m   & \m   & \m
      & \m   &      \\

    \cite{guo_direct_2023}
      & 2023 & Direct training
      & \m   & \x   &
      & \x   & \x   &
      & \m   &      \\

    \cite{khacef_spike-based_2023}
      & 2023 & Local learning
      &      &      & \x
      & \x   &      & \x
      & \m   &      \\

    \cite{rathi_exploring_2023}
      & 2023 & HW design
      & \m   & \m   & \m
      & \m   &      & \x
      &      &      \\

    \cite{frenkel_bottom-up_2023}
      & 2023 & HW design
      & \m   & \m   & \m
      & \m   &      & \x
      & \m   &      \\

    \cite{yiLearningRulesSpiking2023}
      & 2023 & Training survey
      & \x   & \x   & \x
      & \x   & \m   & \m
      & \x   &      \\

    \cite{al_abdul_wahid_survey_2024}
      & 2024 & HW design
      & \m   & \m   & \m
      &      &      & \x
      &      &      \\

    \cite{dampfhoffer_backpropagation-based_2024}
      & 2024 & Training survey
      & \x   & \x   &
      & \x   & \x   & \m
      & \x   &      \\

    \midrule
    \multicolumn{11}{l}{\textit{Scope-driven set}} \\
    \midrule

    \cite{pietrzak_overview_2023}
      & 2023 & Systematic analysis
      & \x   & \x   & \m
      & \x   &      &
      & \m   &      \\

    \cite{eshraghian_training_2023}
      & 2023 & Tutorial / software
      & \m   & \x   & \x
      & \x   &      & \m
      & \m   & \m   \\

    \cite{mishra_survey_2023}
      & 2023 & Cont.\ learning
      & \m   & \m   & \m
      & \m   &      & \m
      &      &      \\

    \cite{wu_review_2024}
      & 2024 & General SNN
      & \m   & \m   & \m
      & \m   & \m   & \m
      & \m   &      \\

    \cite{hu_toward_2024}
      & 2024 & Direct training
      & \x   & \x   & \m
      & \x   & \x   & \m
      & \m   &      \\

    \cite{zhou_direct_2024}
      & 2024 & Direct training
      & \m   & \x   &
      & \x   & \x   & \x
      & \m   &      \\

    \cite{shen_evolutionary_2024}
      & 2024 & Evolutionary
      & \m   & \m   & \m
      & \x   &      &
      & \m   &      \\

    \cite{khan_spiking_2025}
      & 2025 & General SNN
      & \m   & \m   & \m
      & \m   & \m   & \m
      & \m   &      \\

    \cite{ayasi_practical_2025}
      & 2025 & Tutorial / software
      & \m   & \m   & \m
      & \x   &      &
      &      & \m   \\

    \cite{han_progress_2025}
      & 2025 & Direct training
      & \m   & \m   & \m
      & \m   & \m   & \m
      &      &      \\

    \cite{fatima_minhas_continual_2025}
      & 2025 & Cont.\ learning
      & \m   & \m   & \m
      & \m   &      & \m
      &      &      \\

    \cite{mazurekThreeFactorLearningSpiking2025}
      & 2025 & Local learning
      &      &      & \x
      & \x   &      & \m
      & \m   &      \\

    \cite{huo_research_2025}
      & 2025 & Local learning
      &      & \m   & \x
      & \x   &      & \m
      &      &      \\

    \cite{karamimanesh_spiking_2025}
      & 2025 & HW design
      & \m   & \m   & \m
      &      &      & \x
      &      &      \\

    \cite{lin_benchmarking_2025}
      & 2025 & Systematic analysis
      & \m   & \m   & \x
      & \x   &      &
      &      & \x   \\

    \cite{cheng_comprehensive_2025}
      & 2025 & Systematic analysis
      &  \m  & \m   & 
      & \x   &      & \m
      &      & \x   \\

    \midrule
    \textbf{This work}
      & 2026 & Training survey
      & \x   & \x   & \x
      & \x   & \m   & 
      & \x   & \x   \\
    \bottomrule
  \end{tabular}
  \caption{Surveys related to \gls{snn} training, grouped by scope. \emph{Chronology-driven}: earlier surveys, included to trace the evolution of the field. \emph{Scope-driven}: recent surveys differing from this work in focus (hardware, continual learning, large-scale architectures, etc.); local learning rules and cross-method comparisons receive limited coverage in this literature, motivating the present work.
\textbf{Training} (Section~\ref{sec:taxonomy}):
\textit{Conv}~= ANN-to-SNN conversion;
\textit{BP}~= gradient-based direct training;
\textit{Loc}~= local/online rules.
\textbf{Scope}:
\textit{Train}~= training is a primary subject;
\textit{Large}~= focus on large-scale architectures;
\textit{HW}~= neuromorphic hardware design.
\textbf{Methodology}:
\textit{Taxon}~= systematic taxonomy of training methods;
\textit{Bench}~= reproducible cross-method comparisons.
\x~= substantial; \m~= partial coverage; blank~= not a
focus.\label{tab:surveys}}
\end{table*}

%% file: 4-taxonomy.tex
 \section{Taxonomy}
\label{sec:taxonomy}

\begin{figure}
    \centering
    \includegraphics[width=0.9\linewidth]{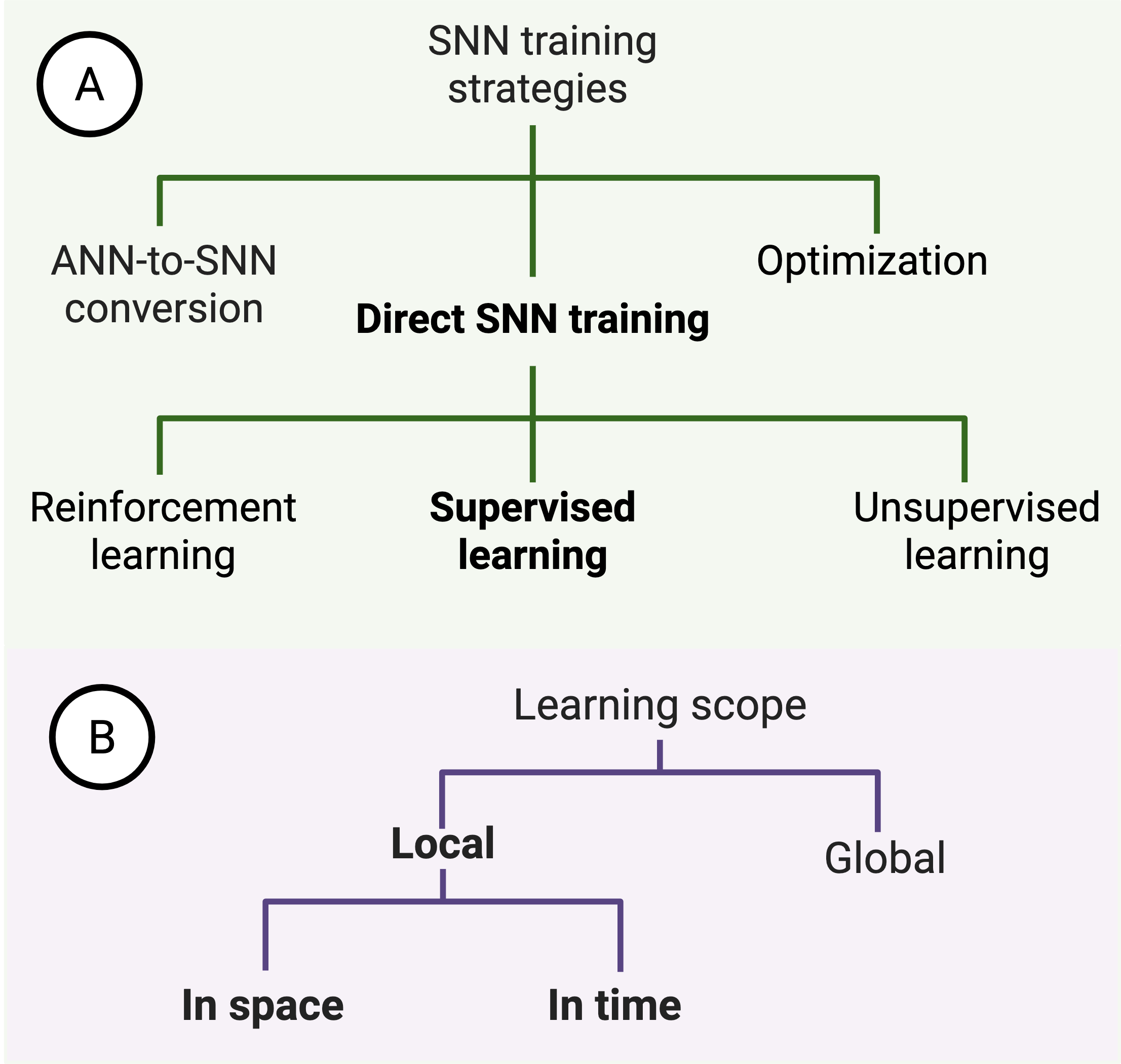}
    \caption{Two complementary taxonomies for SNN training algorithms. (A) Classification by training strategy (ANN-to-SNN conversion, direct training, optimization), with direct training further divided by learning paradigm (reinforcement, supervised, unsupervised). (B) Classification by learning scope (global vs. local), with local methods distinguished by spatial or temporal locality.}
    \label{fig:taxonomy}
\end{figure}

Navigating the literature on training algorithms for \glspl{snn} is increasingly challenging: the number of proposed methods is rapidly growing, and closely related ideas are often presented under different names, with different assumptions or implementation choices. In this section, we introduce a systematic taxonomy, organized along multiple axes, to help readers structure the design space and locate individual contributions. We first describe the taxonomy architecture by defining the categorization axes and motivating their relevance. We then map the current algorithmic landscape onto these axes, justifying the placement of representative works within each category. Figure~\ref{fig:taxonomy} graphically summarizes the complete structure.

\subsection{Taxonomy Categorization Axes}
\label{subsec:taxonomy_axes}

To ease exploration, the taxonomy is organized along a set of categorization axes that reflect the most common characteristics used in the literature to describe and compare \gls{snn} training algorithms. These axes provide a structure for highlighting recurring design choices and positioning different works within a shared framework.

We begin by clarifying what it means to ``train'' an \gls{snn}, as the literature encompasses substantially different approaches under this umbrella. A first key axis is the chosen learning strategy (Figure~\ref{fig:taxonomy}, panel A); we identify three main paradigms: (i) \emph{direct SNN training}, where learning is performed while explicitly simulating spiking dynamics, spanning both gradient-based methods (e.g., surrogate-gradient \gls{bptt}) and non-gradient synaptic plasticity rules (e.g., \gls{stdp}- and three-factor-inspired updates); (ii) \emph{ANN-to-SNN conversion}, where a standard \gls{ann} is first trained with conventional techniques and subsequently mapped to an \gls{snn} for deployment; and (iii) \emph{evolutionary and population-based optimization}, where network parameters, architectures, or both are optimized through black-box search over a population of candidate solutions, generally without computing analytic gradients of the spiking dynamics. Direct training constitutes the largest and most methodologically diverse body of work, introducing the widest range of credit-assignment mechanisms; it therefore forms the primary focus of this survey. Conversion and evolutionary approaches are discussed in dedicated sections to avoid conflating fundamentally different objectives and constraints.

Within direct training strategies, we further organize algorithms based on a second key axis: the type of learning signal available. We distinguish \emph{supervised} algorithms, which leverage labeled targets; \emph{unsupervised} algorithms, which optimize objectives derived from the structure of the input; and \emph{reinforcement learning} algorithms, where learning is driven by reward signals that may be delayed and obtained through interaction with an environment. This axis is central because it directly constrains the form, timing, and observability of credit assignment.

Motivated by the potential computational advantages offered by biological learning mechanisms and by hardware efficiency constraints, contemporary research increasingly emphasizes the importance of locality in training algorithms. This concept is typically decomposed into two distinct dimensions: temporal locality and spatial locality. \emph{Temporal locality} requires that parameter updates at time $t$ depend exclusively on information available at the current timestep, possibly alongside a bounded internal state (e.g., eligibility traces). This contrasts with methods that store and replay complete activity histories across all timesteps, such as the network unrolling performed during the backward pass of \gls{bptt} \cite{werbos_backpropagation_1990}. \emph{Spatial locality} concerns instead the availability of information across the network structure. Consistent with established literature, we define a learning rule as spatially local if it computes synaptic weight changes using only variables physically present at the synapse (i.e., pre- and postsynaptic activity), possibly modulated by a global signal, provided this does not introduce \emph{update locking}. Update locking occurs when weight modifications cannot proceed synchronously with the forward pass but must wait for a global network operation to complete (e.g., a full backward pass), coupling each synaptic update to the network's overall latency.
Based on these definitions, training algorithms can be classified according to their adherence to spatial, temporal, or both locality, or to neither.

In addition to the primary categorization axes described above, we introduce a second layer of categorization based on \emph{cross-cutting algorithmic mechanisms}. While the axes capture high-level properties, many direct-training methods reuse a small set of recurring building blocks that are combined and instantiated in different ways across papers. We therefore use these mechanisms as \emph{non-exclusive descriptors} to annotate algorithms and make shared structure explicit, helping readers identify commonalities and pinpoint the substantive differences between approaches.

From our survey of the literature, we highlight six mechanisms that recur frequently.

\paragraph{Eligibility traces}
A central challenge in training \glspl{snn} is temporal credit assignment: determining how a synaptic weight changes at time $t$ should account for activity that unfolded over preceding timesteps. Many methods address this by equipping each synapse with an \emph{eligibility trace}, an auxiliary state variable that accumulates a temporally decaying record of recent pre- and postsynaptic activity. Rather than storing complete activity histories, traces compress temporal information into a compact, continuously updated quantity that can be maintained locally and forward in time. Weight updates ($\Delta w_{ji}$) are then produced by combining the trace ($e_{ji}$) with a modulatory learning signal ($M$), an error broadcast, a reward, or a target projection, following the general form of a three-factor rule~\cite{fremauxNeuromodulatedSpikeTimingDependentPlasticity2016}:
\begin{equation}
    \Delta w_{ji} \;\propto\; e_{ji} \;\cdot\; M
    \label{eq:three_factor}
\end{equation}

\paragraph{\gls{dfa}}
In standard backpropagation, hidden-layer gradients are computed by transporting the output error backward through the transpose of the forward weight matrices. \emph{\gls{dfa}} \cite{noklandDirectFeedbackAlignment2016} bypasses this by projecting the output error $\mathbf{e}$ directly to each hidden layer $l$ through a fixed random matrix $\mathbf{B}^l$, yielding a learning signal $\mathbf{g}^l = \mathbf{B}^l \mathbf{e}$. This eliminates the weight-transport problem and removes the layer-by-layer backward pass dependency. In the \gls{snn} context, \gls{dfa} is attractive because the error can be broadcast to all layers simultaneously; however, it still requires the output error to be computed, meaning the full forward pass must complete before learning signals become available.

\paragraph{\gls{drtp}}
\emph{\gls{drtp}} \cite{frenkel_learning_2021} replaces the error-based feedback of \gls{dfa} with a direct projection of the target signal to hidden layers: $\mathbf{g}^l = f(\mathbf{B}^l \mathbf{y}^*)$, where $\mathbf{y}^*$ is the target and $f$ a nonlinearity. The crucial distinction is that \gls{drtp} does not require the output error and therefore does not need to wait for the forward pass to complete: the target is known at the beginning of each sample. This makes \gls{drtp} compatible with update-free, fully pipelined training and naturally combines with eligibility traces to produce algorithms that are both temporally and spatially local.

\paragraph{Auxiliary local classifiers}
Rather than relying on a single global loss at the output, some methods attach \emph{auxiliary per-layer structures}, local readout heads, linear classifiers, or custom decoding circuits, to intermediate layers. Each generates a layer-wise loss or teaching signal, enabling weight updates that do not depend on error propagation from the output. This addresses both spatial locality and the vanishing-gradient problem in deep networks, at the cost of additional parameters and the design choice of how to formulate each local objective.

\paragraph{Spatial backpropagation with online temporal credit assignment}
A family of methods retains standard backpropagation across the spatial dimension at each timestep, but decouples temporal credit assignment from the unrolling required by \gls{bptt}. At each time $t$, a conventional backward pass is computed through the layer stack using an instantaneous loss. At the same time, temporal dependencies are handled separately via online mechanisms, such as eligibility traces or filtered presynaptic activity. This achieves constant memory cost with respect to the number of timesteps while preserving the expressive spatial gradients of backpropagation. These methods are not spatially local, but they achieve full temporal locality, offering a practical middle ground between the performance of \gls{bptt} and the constraints of online learning.

\paragraph{STDP-inspired mechanisms}
A substantial body of work builds on the conceptual template of \gls{stdp}, adapting spike-timing-based correlations to supervised objectives, reinforcement signals, or specific coding schemes. While classical \gls{stdp} is a two-factor Hebbian rule driven by relative pre-/post-synaptic spike timing, many modern variants extend it to three-factor form by gating the update with a modulatory signal, bridging unsupervised correlation detection with task-driven learning (cf. Equation~\ref{eq:three_factor}).

Importantly, not all algorithms emphasize any of these mechanisms, and some introduce orthogonal ideas; we therefore treat this list as a pragmatic set of common denominators rather than a complete basis for the full design space.

\subsection{Training Algorithms}
\label{subsec:training_algorithms}
We now examine the core of the taxonomy by analyzing the current state of the art in \gls{snn} training algorithms. In this section, we contextualize the existing literature within the proposed framework and provide the rationale for the structure illustrated in Figure~\ref{fig:taxonomy}. The discussion is organized around the three primary paradigms identified previously: \emph{direct training}, \emph{evolutionary algorithms}, and \emph{ANN-to-SNN conversion}. We prioritize the direct training category, as it represents the most diverse and active area of research and provides a concise overview of conversion and evolutionary approaches. Our analysis of direct training proceeds from supervised learning to unsupervised methods, concluding with reinforcement learning.

\subsubsection{Direct Training}
\label{subsubsec:direct_training}
The direct training paradigm encompasses all strategies that derive parameter updates explicitly from the simulation of \gls{snn} dynamics. This category is the primary focus of our survey, reflecting the rapidly expanding body of work that exploits the temporal richness of spiking neurons to solve complex tasks. Furthermore, direct training arguably offers the most promising path to highly efficient implementations, particularly in low-latency, energy-constrained embedded scenarios.

 \paragraph{Early Supervised Approaches}
The majority of research on \glspl{snn} uses supervised learning because it is a simpler, more stable, and straightforward way to train \glspl{snn}. Supervised learning in traditional neural networks is a mature field, where backpropagation is the most widely used training technique. As a consequence of the supervised learning problem for \glspl{snn}, researchers sought to adapt backpropagation algorithms to the specific dynamics of spiking neurons. 

Early direct supervised methods for \glspl{snn} predominantly relied on explicit coding schemes to circumvent the non-differentiability of spike generation. A prominent paradigm employs exact spike-time (latency) coding, where the loss function is defined directly in terms of output firing times. SpikeProp serves as the seminal example, backpropagating the error between desired and observed spike times by linearizing the membrane potential at threshold crossings \cite{bohte_error-backpropagation_2002}. In its original formulation, this method was tightly coupled to a complex synapse model featuring multiple delayed terminals per connection. This architecture increased parameter complexity, rendering training highly sensitive to initialization and neuronal quiescence. Subsequent variants introduced mechanisms such as delay and threshold adaptation, forced firing for silent units, or accelerated optimization to enhance convergence \cite{schrauwen_extending_2004,mckennochFastModificationsSpikeProp2006,sporea_supervised_2013}. Despite these refinements, the approach faces significant scalability challenges; spike creation and annihilation induce discontinuities in the error landscape, and the requirement for engineered temporal targets limits flexibility.

A second category of methods avoids direct differentiation of spike times, instead comparing output and teacher spike trains through local learning rules. \gls{resume} formulates supervision as the minimization of the discrepancy between desired and generated spike trains, deriving a delta-rule implemented via \gls{stdp}-like correlations \cite{ponulak_supervised_2010}. Similarly, SPAN filters spike trains into continuous traces to enable Widrow-Hoff-style updates in the analog domain \cite{mohemmed_span_2012}. At the same time, the Chronotron optimizes spike-train similarity to reproduce temporal output patterns \cite{florian_chronotron_2012} precisely. Alternatively, the Tempotron targets a binary classification task, updating weights only upon error by utilizing the timing of the maximal membrane potential \cite{gutigTempotronNeuronThat2006}. Finally, SWAT adopts a rate-based coding perspective, encoding inputs as spike trains with class-specific frequencies, combining \gls{stdp} with a \gls{bcm}-like rule for classification \cite{wade_swat_2010}.

Collectively, these methods established the foundational paradigms for supervised spiking learning, including timing-based backpropagation, spike-train matching, and classification-oriented rules. However, their utility in modern contexts is limited; they are typically constrained to shallow architectures (often single layers) and rely on rigid temporal codes. Consequently, they lack the robustness and scalability offered by contemporary surrogate-gradient and \gls{bptt}-based training within standard deep-learning frameworks.

\paragraph{Enabling Backpropagation in SNNs}
Following the initial exploration period, the resounding success of deep \glspl{ann} motivated a paradigm shift toward training \glspl{snn} using gradient-based optimization. The fundamental challenge in this domain is the non-differentiability of the spike generation mechanism $S[t] = \Theta(U[t] - U_{\text{thr}})$. The derivative of the Heaviside step function $\Theta(\cdot)$ is zero almost everywhere and undefined at the threshold, which leads to vanishing (or ill-posed) gradients. It can prevent subthreshold neurons from receiving a useful learning signal. To overcome this, researchers have developed multiple distinct strategies to enable error backpropagation through spiking dynamics.

The strategy pioneered by Lee et al.~\cite{lee_training_2016} circumvented the discontinuity by treating the accumulated membrane potential, rather than the discrete spike, as the primary signal carrying information. By effectively smoothing the signal and ignoring the sharp reset mechanism during the backward pass, this method provided one of the first demonstrations of direct deep \gls{snn} training. However, the resulting mismatch between forward and backward dynamics limited its precision.

A more systematic line of work explicitly adapts \gls{bptt} to spiking dynamics, most notably formalized in \gls{stbp}~\cite{wu_spatio-temporal_2018}. In this view, an \gls{snn} is treated as a discretized \gls{rnn}: the network is unrolled over time and gradients are accumulated jointly along spatial (layer-to-layer) and temporal (time-to-time) dependencies of the membrane state. Practical training is enabled by the \textit{surrogate gradient} trick, which replaces the non-differentiable factor $\partial S[t]/\partial U[t]$ in the backward pass by a bounded proxy $\sigma'(\cdot)$ while keeping the forward pass strictly spiking~\cite{neftci_surrogate_2019}.

Closely related in spirit, but organized around spike-train objectives rather than an explicit unroll-and-backprop implementation, SuperSpike~\cite{zenkeSuperSpikeSupervisedLearning2018} optimizes a spike-train distance (via the van Rossum metric) and introduces a smooth pseudo-derivative in voltage space, yielding an error-modulated learning rule that can be interpreted through eligibility traces and that directly targets precise spike timing.

A complementary ``temporal specialist'' route is exemplified by SLAYER~\cite{shrestha_slayer_2018}, which adopts a Spike Response Model viewpoint and redistributes error through temporal response kernels so that credit assignment aligns with delayed postsynaptic effects; this formulation naturally supports learning not only synaptic weights but also axonal delays on event-driven tasks (e.g., \gls{dvs} gesture recognition \cite{amirLowPowerFully2017}). Building on this line, EXODUS~\cite{bauerEXODUSStableEfficient2023} revisits SLAYER's efficient layer-wise, time-vectorized gradient computation and improves stability by explicitly accounting for the neuron reset in the backward pass via implicit differentiation, reducing the need for ad-hoc gradient scaling while recovering gradients equivalent to \gls{bptt}.

While these approaches highlight different routes to making gradients usable in the presence of spikes, the \gls{stbp}-style surrogate-gradient \gls{bptt} pipeline has arguably become the most widely adopted template in deep-learning-oriented \gls{snn} research, largely because it maps naturally onto modern autodiff frameworks and scales to deep architectures and large datasets. At the same time, the community has clarified that surrogate-gradient training is often surprisingly robust to the \emph{particular} surrogate shape, even though stability can depend strongly on scaling and implementation details~\cite{zenke_remarkable_2021}. Building on this momentum, the field has seen rapid advancements aimed at stabilizing convergence and scaling to deeper architectures.

Key improvements include the introduction of learnable membrane time constants~\cite{fang_incorporating_2021} and unified gated neuron models~\cite{yao_glif_2022} to enhance parametric flexibility. Furthermore, optimization stability has been significantly improved through SNN-specific normalization techniques, including threshold-dependent batch normalization~\cite{zheng_going_2021}, batch normalization through time~\cite{kim_revisiting_2021}, and temporal effective batch normalization~\cite{duan_temporal_2022, guo_membrane_2023}, as well as gradient re-weighting strategies like TET~\cite{deng_temporal_2022}.
Recent works have further refined the learning landscape by rectifying membrane potential distributions~\cite{guo_recdis-snn_2022}, maximizing information flow via novel loss functions~\cite{guo_im-loss_2022}, and enhancing gradient precision via differentiation on spike representations~\cite{meng_training_2022} or learnable surrogate shapes~\cite{lian_learnable_2023}.

Despite these successes, most of these gradient-based methods share a common limitation regarding the proposed taxonomy: they are fundamentally non-local. Spatially, they rely on global error propagation, which violates strict spatial locality. Temporally, explicit \gls{bptt} pipelines require unrolling dynamics over the full simulation horizon and storing intermediate states, leading to a memory footprint of $\mathcal{O}(T)$ and weight updates that are typically locked to the end of the sequence~\cite{eshraghian_training_2023}. Alternative formulations (e.g., trace- or kernel-based credit assignment) can reduce the need to store full trajectories. However, they still depend on temporally extended history and non-local error signals, leaving online learning on memory-limited neuromorphic hardware as a central challenge.

\paragraph{Emergence of locality and bioinspiration}
As surrogate-gradient \gls{bptt} matured into a reliable recipe for high accuracy, its practical drawbacks, update locking, a backward pass spanning the full depth, and the need to retain temporal state for gradient computation, made its mismatch with neuromorphic constraints increasingly apparent \cite{eshraghian_training_2023}. This tension helped revive an older intuition from computational neuroscience: learning in the brain is largely \emph{local}, shaped by pre- and postsynaptic activity and modulated by relatively simple global factors \cite{williams_learning_1989, bi_synaptic_1998, song_competitive_2000}. Early in this renewed direction, several works revisited \gls{stdp} \cite{song_competitive_2000} not merely as an unsupervised feature learner, but as a scaffold for supervision. A representative example is the \emph{stable supervised STDP} lineage \cite{shrestha_stable_2017, goupy_paired_2024}, which keeps weight updates synapse-local by enforcing bounded, weight-dependent potentiation/depression and by injecting a teaching signal at the output so that spike timing becomes class-discriminative. More recent variants such as S2-STDP \cite{goupy_paired_2024} further stabilize learning by coupling supervision to latency objectives (encouraging target neurons to fire early and non-target neurons late) and by structuring the classifier around paired competition, effectively obtaining stronger separability with minimal output circuitry while remaining close to an event-driven, plasticity-rule mindset.

A parallel and highly influential thread sought locality through \emph{eligibility traces}, translating temporal credit assignment into forward-time state variables that resemble synaptic tags; conceptually, this idea traces back to \gls{rtrl}~\cite{williams_learning_1989}, which computes exact online gradients for continually running fully recurrent networks by maintaining state--parameter sensitivities over time. In recurrent settings, \gls{eprop}~\cite{bellec_solution_2020} is emblematic: it factorizes gradients into a product of a synapse-specific eligibility trace and a (possibly broadcast) learning signal, enabling online updates without storing the entire unfolded history, but typically remaining spatially non-local because the learning signal depends on output error and must be delivered to hidden units. This same basic decomposition, a local trace modulated by a teaching term, reappears in many later rules, differing mainly in (i) how closely the trace approximates exact gradients, and (ii) whether the teaching term is obtained by true backpropagation, by random projections, or by fully local objectives. \gls{ostl}~\cite{bohnstingl_online_2023} sharpens this picture by explicitly separating spatial learning signals from temporal traces and clarifying when the resulting updates are gradient-equivalent and when they become controlled approximations (e.g., by dropping higher-order terms to tame complexity). In \cite{yin_accurate_2023} the authors adapt to \glspl{snn} the \gls{fptt} algorithm, which recasts temporal credit assignment as a consensus problem and avoids backward unrolling by augmenting the instantaneous loss with a dynamic regularisation term based on a running average of past weight updates. Combined with liquid time-constant spiking neurons whose input-driven membrane time-constants provide the gating structure \gls{fptt} requires to converge on recurrent tasks.
Several methods then push further along the ``\gls{bptt}-like but cheaper'' axis: \gls{ottt}~\cite{xiao_online_2022} derives instantaneous updates by combining a time-local loss with forward traces, \gls{sltt}~\cite{meng_towards_2023} makes an even more aggressive simplification by discarding temporal gradient paths so that one essentially performs spatial backprop at selected time steps with constant-in-time memory, and STOP \cite{gao_stop_2025} keeps spatial backprop per time step but enriches temporal assignment through traces while extending adaptation beyond weights to include intrinsic parameters such as thresholds and leak factors; S-TLLR \cite{apolinario_s-tllr_2024} applies the same spatial-BP-per-step template with STDP-inspired causal and non-causal eligibility traces. In the same spirit of improving temporal approximations \emph{in deep networks}, OTPE \cite{summe_estimating_2024} highlights that many efficient trace rules under-credit early-layer parameters for errors that only manifest downstream after leakage and synaptic filtering; it restores these delayed postsynaptic influences via additional traces (with a further approximation to recover near-linear scaling). At the end of scalability, ES-D-RTRL/BrainScale \cite{wang_brainscale_2025} revisits the classic \gls{rtrl} objective, exact online gradients, and makes it practical by aggressively approximating the eligibility tensor (e.g., diagonal/rank-structured forms with smoothing), reaching network sizes that are far beyond what synapse-wise trace methods can store.

In contrast to trace-centric approaches that mainly attack \emph{temporal} non-locality, another family targets \emph{spatial} non-locality by removing global backpropagation across layers. \gls{decolle}~\cite{kaiser_synaptic_2020} is a clean example: by attaching auxiliary readouts to each layer and optimizing layer-wise losses, it avoids a global backward pass and allows layers to learn in parallel, using three-factor, surrogate-gradient-style updates driven by local errors and forward traces. A key limitation of strictly fixed/random readouts is ``weak coupling'', early layers may learn representations that satisfy their own local objective but are suboptimal for downstream processing, which motivates refinements where auxiliary classifiers become trainable and where the temporal approximation is made explicit. \gls{ell}/\gls{bell}/\gls{fell}~\cite{ma_deep_2023} fit here: \gls{ell} prioritizes strict efficiency by ignoring temporal dependencies, \gls{bell} performs local \gls{bptt} within each layer to recover accuracy at the cost of time-locking, and \gls{fell} uses forward traces to approximate the temporal component online, offering a middle ground that often narrows the gap to global training while preserving layer-wise decoupling; STDL \cite{ma_spatio-temporal_2025} generalizes the local-classifier idea to multi-layer blocks, using downstream blocks as auxiliary classifiers rather than dedicated readout heads, while discarding temporal gradient terms for fully online updates. A related, hardware-oriented route replaces backpropagated errors with \emph{direct} projections: \gls{stdfa}~\cite{lee_spike-train_2020} adapts direct feedback alignment to spike-train level signals so that hidden layers receive teaching information through fixed random feedback rather than weight-transported gradients; \gls{etlp}~\cite{quintana_etlp_2024} and \gls{osttp}~\cite{ortner_online_2023} push this logic into three-factor, trace-modulated plasticity by combining eligibility traces with \gls{drtp} \cite{frenkel_learning_2021}, aiming for fully online updates with minimal global coordination. \gls{stsf} \cite{he_stsf_2025} takes a particularly pragmatic hybrid stance by pairing a sparse feedback-alignment-style global signal with an \gls{stdp}-like local rule, leveraging sparsity to reduce the cost of feedback while preserving an event-driven learning core.

Finally, a few approaches attempt to avoid the usual backward infrastructure more radically. \gls{tess}~\cite{apolinario_tess_2025} achieves spatial and temporal locality by generating errors locally at each layer through fixed projections into a task space (local error generation), coupled with instantaneous, STDP-flavored traces that include both causal and non-causal components; this keeps learning simple and hardware-friendly, at the price of leaning on carefully designed layer-wise targets. \gls{tp} \cite{pes_traces_2026} goes further by eliminating the backward pass via a forward-only, dual-path scheme in which target information is propagated in parallel to the input-driven activity, enabling synapses to update using only locally available traces and the locally delivered target signal; notably, this forward-only design is among the few fully local rules reported to scale to deep convolutional backbones. A complementary forward-only paradigm draws directly on the \gls{ff} algorithm~\cite{hinton_forward-forward_2022}, which replaces the backward pass with two forward passes over positive (correctly labeled) and negative (incorrectly labeled) data, optimizing a layer-wise \emph{goodness} objective locally. In \cite{ghader_ff_snn_2026} this framework is adapted to \glspl{snn} by using accumulated spike counts as the goodness measure, applying layer normalization between layers to decouple representations across depth, and computing layer-local weight updates via surrogate gradients. Taken together, these works illustrate a broad spectrum of compromises: methods closer to \gls{bptt} tend to preserve accuracy but relax locality (often spatially), while more bio-inspired and hardware-friendly rules improve locality and online capability by introducing auxiliary objectives, random feedback, or forward-only teaching signals, with the central practical challenge remaining how to retain strong performance as depth, temporal structure, and task complexity increase.

Table~\ref{tab:taxonomy_supervised} summarizes the locality and mechanistic 
properties of all supervised direct-training algorithms surveyed above.

\paragraph{Unsupervised approaches for \glspl{snn}}
Unsupervised learning has received sustained attention in \gls{snn} research, motivated by its deep roots in the neuroscience of synaptic plasticity and by the practical appeal of learning rules that require no global error signal, a property well suited to neuromorphic hardware. Although unsupervised methods alone do not yet rival supervised approaches on complex benchmarks, they have produced a diverse methodological landscape and remain essential where labeled data or global feedback pathways are unavailable.

The dominant mechanism underlying unsupervised \gls{snn} training is \gls{stdp}, whose experimental basis was established in the late 90s \cite{markram_regulation_1997, bi_synaptic_1998}: synaptic strength increases when a presynaptic spike precedes a postsynaptic spike, and decreases for the reverse order, with approximately exponential dependence on the temporal difference. In \cite{song_competitive_2000}, \gls{stdp} was shown to function as a competitive Hebbian rule, producing self-organized input selectivity without explicit normalization. Two subsequent works established \gls{stdp} as a viable mechanism for unsupervised representation learning. In \cite{masquelier_unsupervised_2007}, \gls{stdp} in a hierarchical feedforward \gls{snn} with rank-order temporal coding was shown to enable neurons to become selective to repeating spatiotemporal patterns (e.g., faces, object parts) in natural images. Later, 95\% accuracy on MNIST was achieved in \cite{diehl_unsupervised_2015} using a two-layer network of \gls{lif} neurons trained entirely without supervision, demonstrating that combining \gls{stdp} with lateral inhibition, winner-take-all competition, and adaptive thresholds suffices for competitive digit recognition. This work became the standard reference for unsupervised \gls{stdp}-based classification and showed that biologically motivated circuit-level inductive biases can effectively substitute for global optimization.

Classical pair-based \gls{stdp} cannot account for the frequency dependence of plasticity observed experimentally. This was addressed with triplet-based rules~\cite{pfister_triplets_2006,gjorgjievaTripletSpiketimingdependentPlasticity2011}, which extend standard \gls{stdp} with additional traces and connect to the \gls{bcm} learning rule~\cite{bienenstockTheoryDevelopmentNeuron1982}, and with a voltage-based formulation in \cite{clopath_connectivity_2010} where synaptic changes additionally depend on the postsynaptic membrane potential. Network stability requires complementary mechanisms: weight-dependent update rules~\cite{rossumStableHebbianLearning2000, gutigLearningInputCorrelations2003} and homeostatic processes, such as synaptic scaling~\cite{turrigiano_activity-dependent_1998}, maintain weight distributions and firing rates within a functional range. Hardware-oriented simplifications have also been explored, such as \gls{sdsp} with bistable synapses~\cite{brader_learning_2007}, a robust and implementation-friendly alternative to full \gls{stdp}.

Extending \gls{stdp} from shallow to deep architectures is challenging because the rule provides no mechanism for propagating learning signals across layers. The dominant solution is greedy layer-wise training, where each layer is trained with \gls{stdp} independently and then frozen before training the next, analogous to the unsupervised pre-training of deep belief networks~\cite{hintonFastLearningAlgorithm2006}. Since purely unsupervised \glspl{snn} lack a classification mechanism, most works in this area evaluate learned representations through a supervised readout (e.g., SVM or linear probe) appended after training; the feature learning itself, however, remains unsupervised. The first deep convolutional \gls{snn} with multiple \gls{stdp}-trained layers (SDNN) was introduced in~\cite{kheradpisheh_stdp-based_2018}, learning hierarchical features via rank-order temporal coding and achieving 98.4\% on MNIST. Subsequent work addressed the sequential bottleneck of greedy training through a dual-accumulator neuron model enabling simultaneous layer-wise learning from event streams~\cite{thiele_event-based_2018}, showed that \gls{lif} dynamics under rank-order coding reduce to a ReLU-like operation amenable to GPU-accelerated mini-batch \gls{stdp}~\cite{ferreUnsupervisedFeatureLearning2018}, and demonstrated that residual connections are critical for sustaining learning beyond two or three layers~\cite{srinivasanReStoCNetResidualStochastic2019}. Despite these advances, \gls{stdp}-trained networks have been shown to perform effectively only up to roughly five layers, and their accuracy on complex benchmarks such as CIFAR-10 remains well below that of supervised surrogate-gradient methods.

An important theoretical thread provides formal justification for \gls{stdp} as an unsupervised learning principle. In \cite{buesing_neural_2011}, stochastic spiking networks were shown to implement \gls{mcmc} sampling from Boltzmann-like distributions, and in \cite{nessler_bayesian_2013} it was proved that \gls{stdp} in winner-take-all circuits approximates expectation-maximization for mixture models, recasting the rule as an implementation of principled statistical inference rather than an ad hoc biological mechanism. Event-driven contrastive divergence for spiking restricted Boltzmann machines~\cite{neftciEventdrivenContrastiveDivergence2014} and stochastic synaptic sampling~\cite{neftciStochasticSynapsesEnable2016} translated these ideas into engineering practice, with particular relevance for neuromorphic substrates where device-level stochasticity is intrinsic.

More recently, self-supervised paradigms from mainstream deep learning have been adapted to spiking networks. \gls{csdp}~\cite{ororbiaContrastiveSignalDependent2024} is inspired by forward-forward-style local objectives and adapts them to spike-based communication, adjusting synapses locally using contrastive signals without backpropagation. Momentum contrastive learning was adapted to \glspl{snn} in~\cite{maNeuroMoCoNeuromorphicMomentum2024}, achieving strong results on neuromorphic datasets. More broadly, the paradigm of unsupervised \gls{stdp}-based pre-training followed by supervised fine-tuning has also proven effective for deep spiking convolutional networks~\cite{lee_training_2018}. Classical \gls{stdp} has also been revisited with adaptive mechanisms, pushing unsupervised plasticity to CIFAR-10 for the first time~\cite{dongUnsupervisedSTDPbasedSpiking2023}.

Overall, unsupervised methods lie at the local end of the \gls{snn} training spectrum, and their evolution from early \gls{stdp} demonstrations to modern contrastive and generative approaches shows that architectural inductive biases and principled objectives have proven as important as the plasticity rule itself. The scalability ceiling of pure \gls{stdp} remains the central open challenge, increasingly addressed by self-supervised paradigms that provide richer training signals while preserving locality.

\paragraph{Reinforcement learning in \glspl{snn}}
\gls{rl} in \glspl{snn} is grounded in the observation that synaptic plasticity is modulated by neuromodulatory signals (most notably dopamine) whose phasic responses encode reward prediction errors closely matching the temporal-difference signal of \gls{rl} theory~\cite{schultzNeuralSubstratePrediction1997}. This provides a natural instantiation of the three-factor rule (cf.\ Equation~\ref{eq:three_factor}): an \gls{stdp}-derived eligibility trace records local pre- and postsynaptic correlations. At the same time, a global reward signal acts as the modulatory factor that converts the trace into an actual weight change. Unlike supervised methods, the learning signal is scalar, delayed, and evaluative rather than instructive, making credit assignment substantially harder. Two broad algorithmic families have emerged to address this: local reward-modulated rules, which preserve biological plausibility, and surrogate-gradient methods that adapt deep \gls{rl} pipelines to spiking networks.

The theoretical foundations of \gls{rstdp} were established by several concurrent contributions~\cite{florianReinforcementLearningModulation2007,izhikevichSolvingDistalReward2007, farries2007reinforcement}, deriving reward-gated \gls{stdp} updates and showing that transient synaptic eligibility traces can bridge the gap between neural activity and delayed reward signals. Subsequent analysis proved that \gls{rstdp} variants require the modulatory signal to encode reward relative to a learned baseline~\cite{fremauxFunctionalRequirementsRewardModulated2010,legensteinLearningTheoryRewardModulated2008}, motivating spiking actor-critic architectures in which critic neurons supply this baseline via temporal-difference learning~\cite{potjansSpikingNeuralNetwork2009, fremauxReinforcementLearningUsing2013}. A unified taxonomy of spiking \gls{rl} rules (R-max, \gls{rstdp}, and TD-STDP) was provided in~\cite{fremauxNeuromodulatedSpikeTimingDependentPlasticity2016}, which remains the standard theoretical reference for local reward-modulated plasticity.

\gls{rstdp} has since been extended to deeper architectures and applied settings. In~\cite{mozafariFirstSpikeBasedVisualCategorization2018}, reward-modulated \gls{stdp} was shown to train the decision layer of a convolutional \gls{snn} using first-spike temporal coding, with the key observation that \gls{rstdp} extracts task-discriminative features. In contrast, unsupervised \gls{stdp} extracts any recurring pattern. A hybrid strategy combining unsupervised \gls{stdp} in early convolutional layers with \gls{rstdp} in classification layers was demonstrated in~\cite{mozafari_bio-inspired_2019}, and \gls{rstdp} has been applied to end-to-end sensorimotor tasks including lane keeping~\cite{bingEndEndLearning2018} and target-reaching navigation~\cite{bingSupervisedLearningSNN2019}, illustrating the rule's suitability for event-driven robotic control.

\gls{eprop}~\cite{bellec_solution_2020} factorizes loss gradients into synapse-local eligibility traces propagated forward in time and a top-down learning signal available online, fitting exactly the three-factor template. Applied to an evidence-accumulation task requiring integration of cues over hundreds of timesteps, \gls{eprop} with adaptive \gls{lif} neurons succeeded where standard \gls{lif} neurons could not, demonstrating that slow neuronal dynamics are critical for temporal credit assignment under delayed rewards.

More recently, deep \gls{rl} methods have been adapted to train \glspl{snn} using surrogate gradients directly. A pivotal contribution was the \gls{popsan}~\cite{tangDeepReinforcementLearning2021}, which uses population coding at the input and output layers to increase representational capacity. \gls{popsan} trains a spiking actor network alongside a conventional deep critic using standard policy optimization algorithms~\cite{schulman2017proximal, fujimoto2018addressing, haarnoja2018soft}; the critic supplies the policy gradient during training and is discarded at deployment, so only the spiking actor runs on neuromorphic hardware. This hybrid actor-critic design has become the dominant template for deep spiking \gls{rl}: it inherits the training stability of established deep \gls{rl} pipelines while confining the energy-efficient spiking inference to the deployed policy. Deployed on Intel's Loihi neuromorphic chip, \gls{popsan} achieved competitive performance on continuous-control benchmarks \cite{brockman2016openai} with reported energy savings of two orders of magnitude relative to GPU baselines, establishing population coding as a key enabler for scaling spiking \gls{rl} to continuous action spaces. Subsequent work improved on this paradigm: in~\cite{zhangMultiSacleDynamicCoding2022}, a multi-scale dynamic coding scheme outperformed both conventional actor networks and \gls{popsan} on locomotion tasks. In contrast, in~\cite{chenDeepReinforcementLearning2024}, a fully spiking deep Q-network (DSQN) was proposed that uses the membrane voltage of non-spiking output neurons to represent Q-values, achieving competitive performance on 17 Atari games without recourse to an auxiliary \gls{ann} critic. A directly trained spiking Q-network was also demonstrated in~\cite{liuHumanLevelControlDirectly2023}, addressing noisy value estimation inherent in rate-coded spike outputs. The fully spiking actor network with intralayer connections (ILC-SAN)~\cite{chenFullySpikingActor2025} further closed the performance gap with mainstream deep \gls{rl} on continuous control by introducing biologically inspired lateral connectivity within spiking layers. Surrogate-gradient spiking \gls{rl} has also been applied to event-driven robotic settings: in~\cite{zanattaDirectlytrainedSpikingNeural2023} spatio-temporal backpropagation was combined with \gls{rl} for UAV obstacle avoidance using \gls{dvs} inputs, with deployment on a neuromorphic accelerator.

From the perspective of the proposed taxonomy, spiking \gls{rl} methods span a wide locality range: classical \gls{rstdp} rules are both spatially and temporally local and map naturally onto neuromorphic hardware~\cite{mikaitisNeuromodulatedSynapticPlasticity2018}, but are limited to shallow architectures and simple tasks; \gls{eprop} achieves temporal locality while remaining spatially non-local; and surrogate-gradient methods inherit the non-locality of backpropagation while currently achieving the strongest results on complex benchmarks. The central open challenge is closing the performance gap between highly local, hardware-friendly rules and globally trained deep spiking \gls{rl} systems, with spiking world models representing a recent step toward richer model-based settings~\cite{sunSpikingWorldModel2025}.

\subsubsection{ANN-to-SNN Conversion}
\label{subsubsec:ann_to_snn}

ANN-to-SNN conversion sidesteps the core difficulties of direct \gls{snn} training, non-differentiable spikes, temporal credit assignment, and the need for spiking-aware optimization, by offloading learning entirely to a standard \gls{ann} trained with backpropagation and subsequently mapping the resulting parameters to a spiking counterpart. Because the approach inherits the mature toolchain and scalability of conventional deep learning, it has historically achieved some of the highest accuracy on static vision benchmarks. It remains the most accessible route to deploying large pretrained models on neuromorphic hardware.

The theoretical basis is the observation that the firing rate of an integrate-and-fire neuron receiving constant input approximates a ReLU activation: over $T$ timesteps with reset-by-subtraction, the average spike count converges to the analog activation with error $\mathcal{O}(1/T)$~\cite{cao_spiking_2015,rueckauer_conversion_2017}. The standard pipeline trains a deep ReLU-based \gls{ann} (e.g., VGG, ResNet), then replaces each activation with a spiking neuron (typically \gls{lif} or \gls{if}), folds batch-normalization parameters into preceding linear layers, and rescales weights and thresholds so that firing rates remain in a functional range~\cite{diehl_fast-classifying_2015,rueckauer_conversion_2017}. In~\cite{cao_spiking_2015}, the ReLU--IF equivalence was first formalized for convolutional architectures; in~\cite{diehl_fast-classifying_2015} data-driven weight normalization and threshold balancing yielded near-lossless MNIST conversion; and in~\cite{rueckauer_conversion_2017} the framework was extended to handle batch normalization, max-pooling, biases, and softmax, accompanied by a percentile-based threshold calibration that became the de facto standard. Scalability to deep residual architectures and the complete ImageNet benchmark was demonstrated in~\cite{sengupta_going_2019}, although at the cost of up to ${\sim}2{,}500$ simulation timesteps for ResNet architectures.

Two intertwined limitations have driven subsequent research. First, the mismatch between continuous activations and discrete spike counts introduces an \emph{accuracy gap} whose sources were systematically decomposed into clipping error (bounded firing rates cannot represent large activations), quantization error (only $T{+}1$ discrete output levels exist), and unevenness error (asynchronous spike propagation distorts layer-wise input distributions)~\cite{bu_optimal_2023}. Second, rate-coded conversion requires many timesteps for accurate firing-rate estimation, resulting in \emph{high inference latency} that erodes the energy advantage of spiking execution.

Addressing these bottlenecks has proceeded along two complementary axes. On the \emph{low-latency} front, in~\cite{han_rmp-snn_2020} the residual membrane potential (soft-reset) neuron was shown to eliminate the systematic negative bias of hard-reset neurons, reducing the required timesteps from thousands to hundreds. Temporal-switch coding was proposed in~\cite{han_deep_2020} to exploit spike timing rather than rate, achieving an order-of-magnitude reduction in latency. A particularly influential contribution was the Quantization Clip-Floor-Shift (QCFS) activation function~\cite{bu_optimal_2023}, which replaces ReLU during \gls{ann} training so that the source network already produces outputs matching the quantization resolution of $T$ spiking timesteps; this was the first method to achieve competitive conversion at $T{=}4$. Subsequent work pushed latency further through optimized membrane potential initialization~\cite{bu_optimized_2022}, iterative retraining down to a single timestep~\cite{chowdhury_towards_2022}, and conversion from quantized \glspl{ann} that formally establishes the equivalence between $T$-timestep spiking inference and ${\sim}\log_2(T)$-bit weight quantization~\cite{hu_fast-snn_2023}. On the \emph{calibration} axis, post-training correction methods adjust thresholds and other parameters using only a small calibration set (often ${\sim}128$ images), avoiding full retraining. In~\cite{li_free_2021}, layer-wise activation transplanting was introduced, enabling conversion of efficient architectures such as MobileNet for the first time; in~\cite{deng_optimal_2021}, a shift-threshold-ReLU formulation provided analytically grounded bias correction; and in~\cite{li_error-aware_2024}, a second-order error analysis extended calibration to object detection on MS~COCO with privacy-preserving synthetic data.

A third line of work blurs the boundary between conversion and direct training. Hybrid schemes use converted weights as initialization and then fine-tune with surrogate-gradient backpropagation, jointly optimizing membrane leak, firing threshold, and synaptic weights to achieve strong accuracy at very low latency~\cite{rathi_diet-snn_2023}; progressive tandem learning~\cite{wu_progressive_2022} and optimized few-spike neuron models~\cite{stockl_optimized_2021} similarly combine the strengths of both paradigms. A conversion to a detection architecture was first demonstrated with Spiking-YOLO~\cite{kim_spiking-yolo_2020}, which introduced channel-wise normalization and signed neurons to map Leaky-ReLU activations. Most recently, the frontier has shifted to Transformer architectures, where softmax, layer normalization, and multiplicative attention lack natural spiking equivalents. In~\cite{you_spikezip-tf_2024}, exact mathematical equivalence between quantized Transformers and converted \glspl{snn} was established through spike-based softmax and layer-normalization operators, and in~\cite{hwang_spikedattention_2024}, training-free Transformer-to-SNN conversion was achieved for both vision and NLP models with minimal accuracy degradation.

All learning occurs in the \gls{ann} domain via dense backpropagated gradients, and the \gls{snn} is obtained through a separate mapping and optional calibration stage. This makes conversion highly practical, it is the easiest route to deploying existing pretrained models on neuromorphic hardware, and it has delivered some of the highest reported accuracies on static vision benchmarks. However, it carries fundamental limitations that motivate the emphasis on direct training in this survey. Conversion inherently decouples learning from spiking dynamics: the network never trains on spike-based representations, limiting its ability to exploit temporal coding, adapt to event-driven inputs, or perform on-chip learning. Even state-of-the-art low-latency methods typically rely on rate coding and require strict architectural constraints (e.g., ReLU-only activations, folded normalization), and the accuracy gap, while dramatically reduced, tends to widen on tasks that require fine temporal structure, such as neuromorphic event-stream processing. Furthermore, conversion offers no path toward on-device adaptation, a central goal for neuromorphic systems operating in non-stationary environments. We therefore treat conversion primarily as an important reference point and upper-bound baseline when evaluating the direct, local training methods that form the core of this work.

\subsubsection{Optimization}
Gradient-free methods optimize \gls{snn} parameters by treating the network as a black box evaluated against a fitness function, entirely sidestepping the non-differentiability of spiking dynamics. Credit is assigned through population-level fitness comparisons or global perturbation--reward correlations rather than through local synaptic quantities.

\paragraph{Evolutionary parameter and topology optimization}
Evolutionary algorithms were among the earliest methods applied to \gls{snn} training. Genetic algorithms were used in~\cite{floreano2001evolution} to evolve synaptic weights of spiking controllers for robot navigation, and differential evolution was shown in~\cite{pavlidisSpikingNeuralNetwork2005} to solve supervised classification tasks with compact spiking architectures. Multi-objective evolutionary search was applied in~\cite{jinEvolutionaryMultiobjectiveOptimization2007} to co-optimize accuracy, weights, delays, and connectivity via Pareto-based selection. A key strength of these methods is their ability to jointly optimize discrete and continuous parameters, connectivity masks, axonal delays, firing thresholds, that gradient methods cannot handle natively. When the goal extends to discovering network structure, NEAT-based neuroevolution has been adapted to spiking neurons for control tasks~\cite{qiuEvolvingSpikingNeural2018} and deployed on neuromorphic hardware~\cite{vandesompeleNeuroevolutionSpikingNeural2016}, while the EONS framework~\cite{schumanEvolutionaryOptimizationNeuromorphic2020} provides general-purpose co-evolution of topology, weights, and neuron-level parameters for neuromorphic deployment. More recently, evolution strategies have been applied to evolve sparse connectivity distributions for recurrent \glspl{snn}~\cite{wangEvolvingConnectivityRecurrent2023}, and differential evolution with cosine-annealing schedules has been used to train deeper architectures on CIFAR-10/100~\cite{jiangCADECosineAnnealing2024}. Neural architecture search extends these ideas to structured, cell-based search spaces: approaches such as AutoSNN~\cite{naAutoSNNEnergyEfficientSpiking2022} and SpikeDHS~\cite{cheDifferentiableHierarchicalSurrogate2022} use evolutionary or differentiable outer search. However, their inner-loop weight training typically relies on surrogate-gradient \gls{bptt}, making their taxonomic status ambiguous. A comprehensive survey of evolutionary methods for \glspl{snn} is provided in~\cite{shen_evolutionary_2024}.

\paragraph{Perturbation-based and other gradient-free methods}
An intermediate class of methods estimates gradients through stochastic perturbation. The foundational work of~\cite{seungLearningSpikingNeural2003} proposed learning by correlating stochastic synaptic noise with a global reward signal, and~\cite{fieteGradientLearningSpiking2006} extended this to estimate gradients by dynamically perturbing membrane conductances. More recently, LocalZO~\cite{mukhotyDirectTrainingSNN2023} proposed neuron-level zeroth-order perturbation with a formal connection to surrogate gradients, achieving competitive accuracy on CIFAR-10/100 and neuromorphic datasets with training speedups over \gls{bptt}. Beyond evolutionary and perturbation-based approaches, particle swarm optimization~\cite{vazquezTrainingSpikingNeurons2011} and other metaheuristics~\cite{javanshirTrainingSpikingNeural2023} have been applied to \gls{snn} weight tuning, though exclusively on small-scale tasks.

\paragraph{Limitations and outlook}
Despite their flexibility, no purely gradient-free method has yet matched the performance of surrogate-gradient \gls{bptt} on large-scale benchmarks such as ImageNet. The most competitive large-scale results come from hybrid \gls{nas} approaches that confine evolutionary search to the outer architectural loop while relying on gradient-based weight training internally. The genuine value of non-standard optimization lies in hardware-constraint-aware search, multi-objective optimization over non-differentiable quantities (energy, latency, spike count), and compatibility with arbitrary neuron models and mixed discrete--continuous parameter spaces. As neuromorphic deployment matures and hardware constraints become first-class optimization objectives, the role of these methods is poised to grow.

\begin{table*}[p]
\centering
\scriptsize
\setlength{\tabcolsep}{3pt}
\renewcommand{\arraystretch}{0.95}
\resizebox{\textwidth}{!}{%
\begin{tabular}{@{} l r c c c c c c c @{}}
\toprule
Algorithm & Year &
  \shortstack{Local\\in time} &
  \shortstack{Local\\in space} &
  Traces &
  \shortstack{STDP-\\inspired} &
  \shortstack{Spatial BP +\\online temp.} &
  \shortstack{Feedback\\align.} &
  \shortstack{Local\\class./readout} \\
\midrule
\multicolumn{9}{@{}l}{\textit{Early approaches}} \\[2pt]
SpikeProp~\cite{bohte_error-backpropagation_2002}         & 2002 &   &   &   &   &   &      &   \\
Tempotron~\cite{gutigTempotronNeuronThat2006}             & 2006 &   & \m &   &   &   &      &   \\
ReSuMe~\cite{ponulak_supervised_2010}                     & 2010 & X & X & X & X &   &      &   \\
SWAT~\cite{wade_swat_2010}                                & 2010 &   &   & X & X &   &      &   \\
SPAN~\cite{mohemmed_span_2012}                            & 2012 &   & \m & X &   &   &      &   \\
Chronotron~\cite{florian_chronotron_2012}                 & 2012 &   & \m &   &   &   &      &   \\
\midrule
\multicolumn{9}{@{}l}{\textit{Enabling backpropagation in SNNs}} \\[2pt]
Lee~et~al.~\cite{lee_training_2016}                       & 2016 &   &   &   &   &   &      &   \\
STBP~\cite{wu_spatio-temporal_2018}                       & 2018 &   &   &   &   &   &      &   \\
SuperSpike~\cite{zenkeSuperSpikeSupervisedLearning2018}   & 2018 & X &   & X &   &   & \m   &   \\
SLAYER~\cite{shrestha_slayer_2018}                        & 2018 &   &   &   &   &   &      &   \\
EXODUS~\cite{bauerEXODUSStableEfficient2023}              & 2023 &   &   &   &   &   &      &   \\
\midrule
\multicolumn{9}{@{}l}{\textit{Trace-centric \& spatial BP with online temporal credit}} \\[2pt]
E-prop~\cite{bellec_solution_2020}                        & 2020 & X &   & X &   &   &      &   \\
OTTT~\cite{xiao_online_2022}                              & 2022 & X &   & X &   & X &      &   \\
OSTL~\cite{bohnstingl_online_2023}                        & 2023 & X &   & X &   & X &      &   \\
SLTT~\cite{meng_towards_2023}                             & 2023 & X &   &   &   & X &      &   \\
FPTT~\cite{yin_accurate_2023}                             & 2023 & X &   &   &   &   &      &   \\
S-TLLR~\cite{apolinario_s-tllr_2024}                      & 2024 & X &   & X & X & X &      &   \\
OTPE~\cite{summe_estimating_2024}                         & 2024 & X &   & X &   & X &      &   \\
STOP~\cite{gao_stop_2025}                                 & 2025 & X &   & X &   & X &      &   \\
ES-D-RTRL / BrainScale~\cite{wang_brainscale_2025}        & 2025 & X &   & X &   &   &      &   \\
\midrule
\multicolumn{9}{@{}l}{\textit{STDP-based supervised locality}} \\[2pt]
SSTDP~\cite{shrestha_stable_2017}                         & 2017 & X & X &   & X &   &      &   \\
S2-STDP~\cite{goupy_paired_2024}                          & 2024 & X & X &   & X &   &      & X \\
\midrule
\multicolumn{9}{@{}l}{\textit{Spatial locality via local classifiers}} \\[2pt]
DECOLLE~\cite{kaiser_synaptic_2020}                       & 2020 & X & X & X &   &   &      & X \\
ELL~\cite{ma_deep_2023}                                   & 2023 & X & X &   &   &   &      & X \\
BELL~\cite{ma_deep_2023}                                  & 2023 &   & X &   &   &   &      & X \\
FELL~\cite{ma_deep_2023}                                  & 2023 & X & X & X &   &   &      & X \\
STDL~\cite{ma_spatio-temporal_2025}                        & 2025 & X & X &   &   & \m &      & X \\
FF-SNN~\cite{ghader_ff_snn_2026}                           & 2026 &   & X &   &   &   &      & X \\
\midrule
\multicolumn{9}{@{}l}{\textit{Spatial locality via feedback alignment}} \\[2pt]
ST-DFA~\cite{lee_spike-train_2020}                        & 2020 &   &   &   &   &   & DFA  &   \\
OSTTP~\cite{ortner_online_2023}                           & 2023 & X & X & X &   &   & DRTP &   \\
ETLP~\cite{quintana_etlp_2024}                            & 2024 & X & X & X &   &   & DRTP &   \\
STSF~\cite{he_stsf_2025}                                  & 2025 & X &   & X & X &   & DFA  & \m \\
\midrule
\multicolumn{9}{@{}l}{\textit{Fully local approaches}} \\[2pt]
TESS~\cite{apolinario_tess_2025}                           & 2025 & X & X & X & X &   &      & X \\
TP~\cite{pes_traces_2026}                                  & 2026 & X & X & X &   &   &      &   \\
\bottomrule
\end{tabular}%
}
\caption{Taxonomy of \textbf{Supervised} direct-training algorithms for \glspl{snn}. \textbf{Local in time / space}: adherence to temporal or spatial locality. \textbf{Traces}: uses eligibility traces. \textbf{STDP-inspired}: learning rule inspired by spike-timing-dependent plasticity. \textbf{Spatial BP + online temp.}: retains spatial backpropagation per timestep while handling temporal credit assignment online. \textbf{Feedback align.}: uses direct feedback alignment (DFA) or direct random target projection (DRTP). \textbf{Local class./readout}: employs auxiliary per-layer classifiers or readout heads. X~=~present, \m ~=~partial/approximate, empty~=~absent.}
\label{tab:taxonomy_supervised}
\end{table*}

%% file: 5-software.tex
\section{NeuroTrain framework}
\label{sec:software_framework}

As discussed in the previous sections, the rapid emergence of new training algorithms in the neuromorphic computing community makes the field increasingly difficult to track and, more importantly, makes fair comparison challenging. Training algorithms are often introduced in separate codebases with heterogeneous datasets, preprocessing pipelines, network architectures, optimization procedures, and reporting conventions. As a result, it is often difficult to separate the actual contribution of a training algorithm from the particular experimental choices used to evaluate it.

\begin{figure*}[!t]
    \centering
    \includegraphics[width=0.8\linewidth]{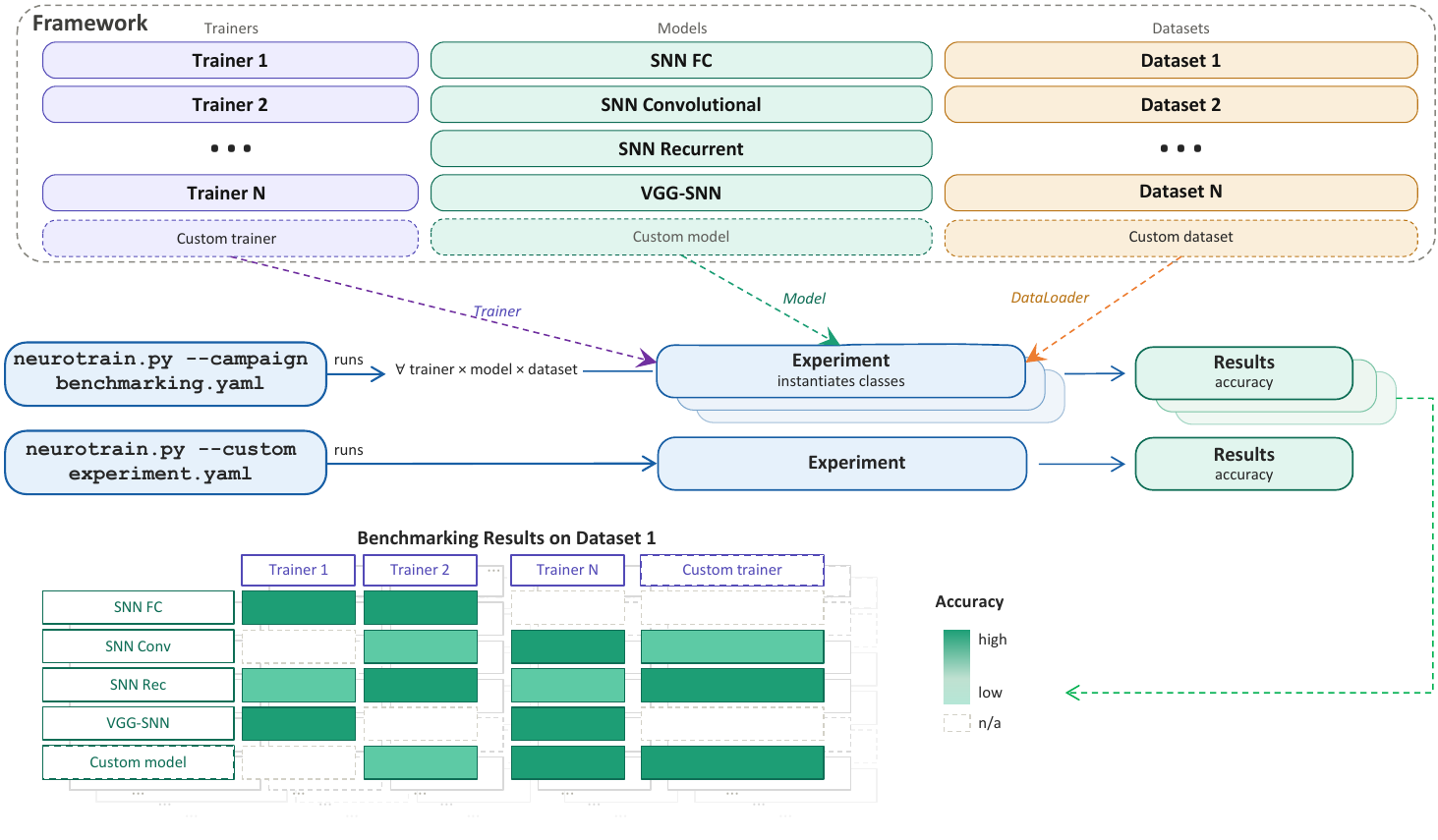}
    \caption{NeuroTrain framework and benchmarking workflow.
The framework organizes SNN training experiments around modular trainers, models, and datasets, including user-defined extensions. A campaign configuration automatically runs all trainer–model–dataset combinations, while a custom configuration executes a single experiment. Results are collected as accuracy scores and summarized in benchmark matrices for systematic comparison across methods. }
    \label{fig:framework}
\end{figure*}

A similar challenge has recently been addressed at the broader level of neuromorphic computing by NeuroBench \cite{yik2025neurobench}, a community-driven benchmarking framework designed to support fair and consistent evaluation of neuromorphic methods and systems, with particular focus on shared tasks, datasets, metrics, and software tools. Inspired by the same need for transparency, reproducibility, and comparability, we introduce \emph{NeuroTrain}, an open framework specifically tailored for implementing, comparing, and benchmarking \gls{snn} training algorithms. NeuroTrain is conceived as a community-oriented resource where researchers can contribute implementations of training algorithms within a shared, standardized environment. In its current version, the framework includes a set of representative state-of-the-art learning rules. Its objective, however, is not to declare a universally best rule, but rather to provide a common setup in which different approaches can be implemented, analyzed, and compared under consistent conditions.
\autoref{fig:framework} illustrates the overall organization of NeuroTrain.

The framework is designed around two main principles: \emph{modularity} and \emph{compartmentalization}. Modularity makes NeuroTrain easily extensible, allowing new datasets, models, and learning rules to be added with limited effort. Compartmentalization keeps the different components isolated, reducing unintended interactions and experimental artifacts that may bias results. Most importantly, this design promotes orthogonality across the framework's elements, enabling systematic combinations of datasets, models, and trainers whenever supported by the corresponding learning rule.
Among the \gls{snn} software frameworks currently available, including snnTorch \cite{eshraghian_training_2023}, SLAYER \cite{shrestha_slayer_2018}, and SpikingJelly \cite{fang2023spikingjelly}, we chose to build NeuroTrain on top of snnTorch. This choice is motivated by its broad adoption in the research community, active maintenance, and tight integration with PyTorch, which provides a flexible and robust software foundation for implementing and benchmarking learning algorithms.

Neurotrain foundation comprises three main building blocks realized as abstract Python classes: (i) \emph{dataloaders}, (ii) \emph{models}, and (iii) \emph{trainers}.

\subsection{Dataloaders}
Since benchmarking training algorithms necessarily depends on the availability of datasets, NeuroTrain begins from this component. To avoid overlap with initiatives such as NeuroBench, NeuroTrain does not aim to define a new dataset suite. Instead, it provides a standardized dataloader interface that allows developers to uniformly integrate datasets into the framework. The current version includes dataloaders for several widely used neuromorphic and rate-encoded datasets, including those available through NeuroBench \cite{yik2025neurobench} and Tonic \cite{lenz2021tonic}, while preserving the ability to add new ones in the future.

\subsection{Models}
Network models are particularly important when benchmarking training algorithms. Although many papers refer to common architectures such as VGG-style models or standard feedforward and recurrent networks, their actual implementations often differ in subtle but important ways and are frequently tightly coupled to the learning rule under study. While such tight integration may be effective for demonstrating the best-case performance of a specific algorithm, it complicates fair comparison. For this reason, NeuroTrain adopts a different approach: it enforces a clear separation between learning rules and network models, provides guidelines for representing networks within the framework, and includes a library of benchmark models. The current library spans fully connected feedforward networks, recurrent models, deep architectures, pure \gls{lif}-based networks, and mixed convolutional models. At present, all benchmark networks rely on the standard \gls{lif} neuron model available in snnTorch, although the framework can be extended with custom neuron models when needed.

\subsection{Trainers}
The third pillar of NeuroTrain is the set of \emph{trainers}. Trainers are Python classes integrated within the snnTorch environment that implement specific learning rules. They follow a common interface so that, whenever compatible, they can operate directly with all supported dataloaders and network models. This provides the basis for orthogonal benchmarking of learning rules across shared tasks and architectures. Naturally, this orthogonality has practical limits: some rules may not apply to certain model classes, such as deep convolutional architectures, and therefore cannot be evaluated on the corresponding datasets. Nevertheless, the common trainer abstraction substantially improves the consistency and breadth of comparisons.

\subsection{Benchmarking framework}

In addition to its three main components, NeuroTrain provides a \emph{benchmarking engine} that automates the execution of experiments generated by systematically combining trainers, models, and datasets, while collecting relevant evaluation metrics. The compartmentalized design of the framework ensures that different learning algorithms are executed within the same software environment and under consistent experimental conditions, thereby enabling fairer, more reproducible comparisons.

NeuroTrain supports two main execution modes: \emph{campaign} and \emph{custom}. In campaign mode, the framework automatically generates an orthogonal experimental campaign by combining selected dataloaders, models, and trainers. During campaign generation, NeuroTrain automatically excludes invalid configurations whenever incompatibilities arise between a trainer and a model, or between a model and a dataset. To preserve encapsulation, such incompatibilities are defined within the corresponding objects, namely trainers, dataloaders, and models, and are made available to the NeuroTrain engine during experimental setup.
We acknowledge that this form of standardization may not always reproduce the absolute best performance achievable by each algorithm, since some methods may benefit from customized implementations, specialized architectures, or highly tuned experimental settings. However, the broader objective of NeuroTrain is to move beyond isolated, algorithm-specific evaluation scripts toward an open benchmarking ecosystem for the \gls{snn} community, similar to the role shared libraries and benchmarks play in other areas of artificial intelligence.

To address the need for more targeted evaluations, NeuroTrain also provides a custom execution mode. This mode allows users to define specific experiments to answer particular research questions or to stress-test a given training algorithm, model, or dataset. A custom experiment is defined as a predefined combination of a trainer, a model, and a dataset, possibly accompanied by user-specified hyperparameters and execution settings.

To further improve fairness, NeuroTrain integrates \emph{Optuna} \cite{akiba2019optuna} for hyperparameter optimization. This allows each training algorithm to be evaluated using a tuned set of hyperparameters rather than arbitrary fixed settings. The framework reports a range of metrics relevant to training and benchmarking, including training accuracy, test accuracy, loss, total execution time, and time per epoch. In addition, for each network, NeuroTrain reports structural and efficiency-related quantities, such as the number of parameters, memory footprint, and activation sparsity.

To demonstrate the framework's capabilities, the current release of NeuroTrain includes implementations of a representative set of learning rules surveyed in this paper. These implementations reflect our effort to interpret the corresponding papers and, whenever possible, to adapt existing code to NeuroTrain's design principles. Consequently, some implementations may still be improved, refined, or extended. NeuroTrain is therefore intended as an open and community-driven project, and contributions from the community are strongly encouraged. In this spirit, the set of implemented training algorithms, models, and datasets is expected to evolve over time, supporting the continued development of the research community.

\begin{figure*}[!t]
    \centering
    \includegraphics[width=\linewidth]{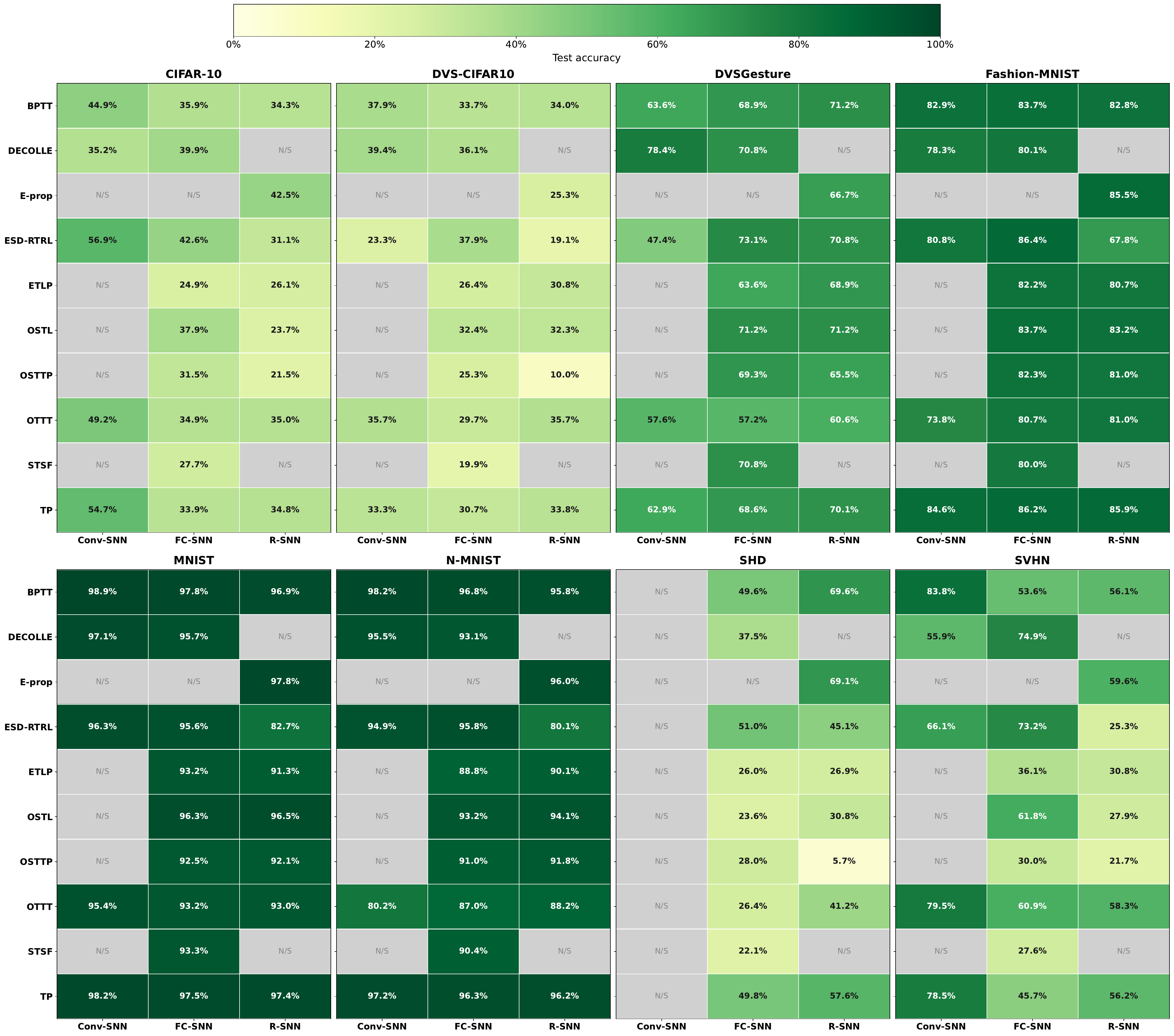}
    \caption{Example of benchmarking results generated by NeuroTrain. Three architectures are benchmarked in variants tailored to the dataset. In the order MNIST, F-MNIST, CIFAR-10, SVHN, N-MNIST, DVS Gesture, DVS CIFAR-10, SHD: \textbf{FC} (784-256-10 / 784-800-10 / 3072-1024-512-10 / 3072-1024-512-10 / 2312-512-10 / 32768-2048-11 / 32768-1024-512-10 / 700-512-20), \textbf{RC} (784-256-10 / 784-256-10 / 3072-512-256-10 / 3072-512-256-10 / 2312-256-10 / 32768-1024-11 / 32768-512-10 / 700-512-20), and \textbf{Conv} (12C5-MP2-32C5-MP2-FC, fixed across datasets). Eprop replaces all multi-hidden-layer architectures with a single-hidden-layer recurrent SNN with 512 units. Each experiment was run for 20 epochs, and hyperparameter exploration was performed using Optuna with 10 trials per experiment for a total of about 850 GPU hours. }
    \label{fig:fresults}
\end{figure*}

To showcase NeuroTrain's benchmarking capabilities, \autoref{fig:fresults} presents a representative snapshot of the results available at the time of submission. The snapshot refers to a benchmarking campaign evaluating the available trainers across a set of small-scale network models with different layer configurations and connectivity patterns. Each experiment was run for 20 epochs, and hyperparameter exploration was performed using Optuna with 10 trials per experiment. Further details on the experimental setup are provided in the official GitHub repository (see \autoref{sec:data-avail}). Overall, the campaign required approximately 850 GPU hours.

The heatmap shows the test accuracy for each experiment. Entries marked as N/S indicate configurations that are not supported by the corresponding trainer or model. The figure illustrates how NeuroTrain enables systematic comparison across trainers, network architectures, and tasks, thereby providing a compact overview of their interactions. These results are not intended to establish a definitive ranking of algorithms. Rather, they demonstrate how NeuroTrain can support exploratory benchmarking and help identify research gaps where further methodological improvements may be needed.

Because NeuroTrain is designed as a living, open resource that will continue to evolve through new implementations, updated configurations, and additional benchmark results, we deliberately avoid embedding the complete benchmark outcomes in this paper. Instead, readers are referred to the official GitHub repository, where the most recent implementations, configurations, and benchmarking results will be maintained.

%% file: 6-conclusion.tex
\section{Conclusions and Future Challenges}
\label{sec:conclusion}

This paper surveyed the rapidly evolving landscape of training algorithms for spiking neural networks, highlighting both the diversity of existing approaches and the difficulty of comparing them fairly across heterogeneous experimental settings. Beyond reviewing the main methodological families, we introduced NeuroTrain, an open framework designed to support the implementation, benchmarking, and comparison of \gls{snn} training rules within a standardized, common environment. By decoupling datasets, network models, training rules, and benchmarking procedures, NeuroTrain aims to improve transparency, reproducibility, and accessibility in this research area, while fostering community-driven development of shared tools and implementations.

As a takeaway, several important challenges remain open. First, many training rules still exhibit limited portability across datasets and architectures, making it difficult to assess their true generality. Second, fair benchmarking requires not only common software interfaces but also careful treatment of hyperparameter optimization, computational cost, memory usage, sparsity, and scalability. Third, the field would benefit from stronger convergence between algorithmic benchmarking and hardware-aware evaluation, especially as neuromorphic systems move toward real-world deployment. Future work will therefore focus on extending NeuroTrain with additional training rules, broader dataset and model coverage, richer benchmarking metrics, and tighter integration with neuromorphic benchmarking initiatives and hardware-oriented evaluation flows. In this perspective, NeuroTrain is intended not as a final benchmark, but as an evolving foundation for more systematic and reproducible progress in SNN training research.

\section*{Data Availability and use of AI \label{sec:data-avail}}

The NeuroTrain framework and its associated implementations are publicly available at the official GitHub repository: \url{https://github.com/smilies-polito/neurotrain}. 

Generative AI tools were used to support language editing, proofreading, and the improvement of writing clarity. They were also used to assist in identifying potentially relevant literature. All suggested references were subsequently verified, selected, and analyzed by the authors. The scientific content, interpretation of the literature, and final manuscript remain the sole responsibility of the authors.